\documentclass{article}

\usepackage[accepted,nohyperref]{icml2021}
\usepackage[ref]{leaf}

\newcommand{\uvec}{\mathbf{u}}
\newcommand{\vvec}{\mathbf{v}}
\newcommand{\zvec}{\mathbf{z}}
\newcommand{\mvec}{\mathbf{m}}
\newcommand{\Avec}{\mathbf{A}}
\newcommand{\Kvec}{\mathbf{K}}
\newcommand{\Zvec}{\mathbf{Z}}
\newcommand{\Svec}{\mathbf{\Sigma}}
\newcommand{\hpvec}{\mathbf{\theta}}

\newcommand{\tikzcircle}[2][red,fill=red]{\tikz[baseline=-0.5ex]\draw[#1,radius=#2] (0,0) circle ;}%

\newcommand{\codeurl}{\href{https://u.perhapsbay.es/vargp-code}{\texttt{u.perhapsbay.es/vargp-code}}}

\icmltitlerunning{Variational Auto-Regressive Gaussian Processes for Continual Learning (VAR-GPs)}

\begin{document}

\twocolumn[
\icmltitle{Variational Auto-Regressive Gaussian Processes for Continual Learning}

\icmlsetsymbol{equal}{*}

\begin{icmlauthorlist}
\icmlauthor{Sanyam Kapoor}{nyu}
\icmlauthor{Theofanis Karaletsos}{fb}
\icmlauthor{Thang D. Bui}{usyd}
\end{icmlauthorlist}

\icmlaffiliation{nyu}{Center for Data Science, New York University, New York, NY, USA}
\icmlaffiliation{fb}{Facebook Inc., Menlo Park, CA, USA}
\icmlaffiliation{usyd}{University of Sydney, Sydney, NSW, Australia}

\icmlcorrespondingauthor{Sanyam Kapoor}{sanyam@nyu.edu}

\icmlkeywords{Gaussian processes, continual learning, variational inference}

\vskip 0.3in
]

\printAffiliationsAndNotice{Work done when the authors were at Uber AI, San Francisco, CA, USA.}

\begin{abstract}
Through sequential construction of posteriors on observing data online, Bayes' 
theorem provides a natural framework for continual learning. We develop 
\emph{Variational Auto-Regressive Gaussian Processes} (VAR-GPs), a principled 
posterior updating mechanism to solve sequential tasks in continual learning. 
By relying on sparse inducing point approximations for scalable posteriors, we 
propose a novel auto-regressive variational distribution which reveals two 
fruitful connections to existing results in Bayesian inference, expectation 
propagation and orthogonal inducing points. Mean predictive entropy estimates 
show VAR-GPs prevent catastrophic forgetting, which is empirically supported by 
strong performance on modern continual learning benchmarks against competitive 
baselines. A thorough ablation study demonstrates the efficacy of our modeling 
choices.
\end{abstract}

\section{Introduction}

Continual Learning (CL) is the constant development of complex behaviors by 
building upon previously acquired skills 
\citep{ring1994continual,thrun1998lifelong}; humans and other animals exhibit 
knowledge acquisition for continual skill development 
\citep{hoppitt2013social}. To the contrary, modern artificial intelligence 
methods based on supervised machine learning rely on a stronger assumption of 
all representative information being available at once, i.e.~\emph{i.i.d} data. 
Many systems, however, violate this assumption. A hospital may not have legal 
access to past patient data to provide automated diagnosis, real-time learning 
systems may be limited by compute to utilize all available data, and a mobile 
device may prefer on-device learning and inference to guarantee user privacy. 
In each of these examples, the learning algorithm does not observe \emph{i.i.d} 
data, but only new parts of the data space. We still want to improve our models 
using new data without compromising existing performance.

\begin{figure}[!t]
\centering
\begin{tabular}{c}
\includegraphics[width=.95\linewidth]{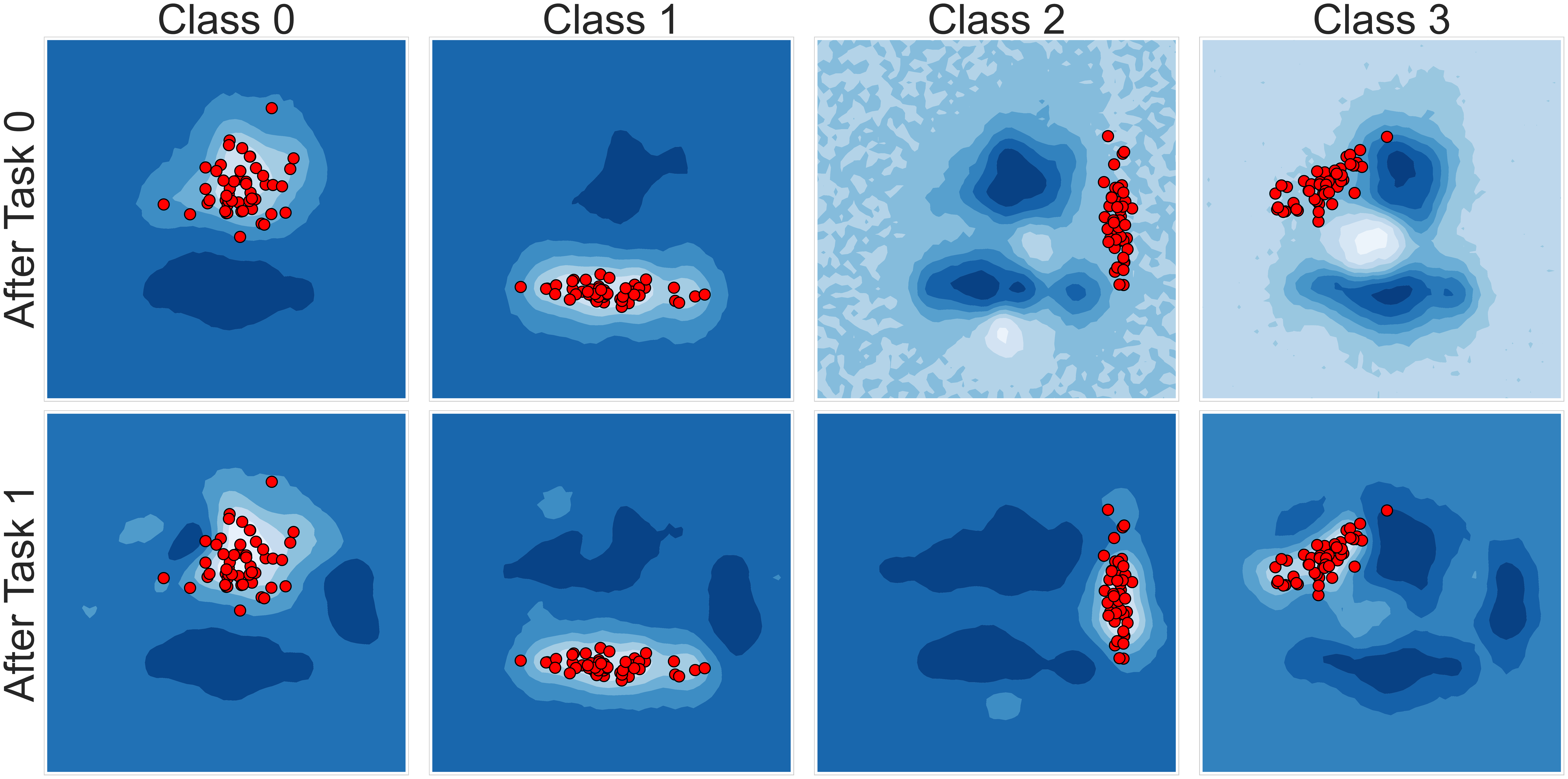} \\ 
(a) VAR-GP (ours) \\ \\
\includegraphics[width=.95\linewidth]{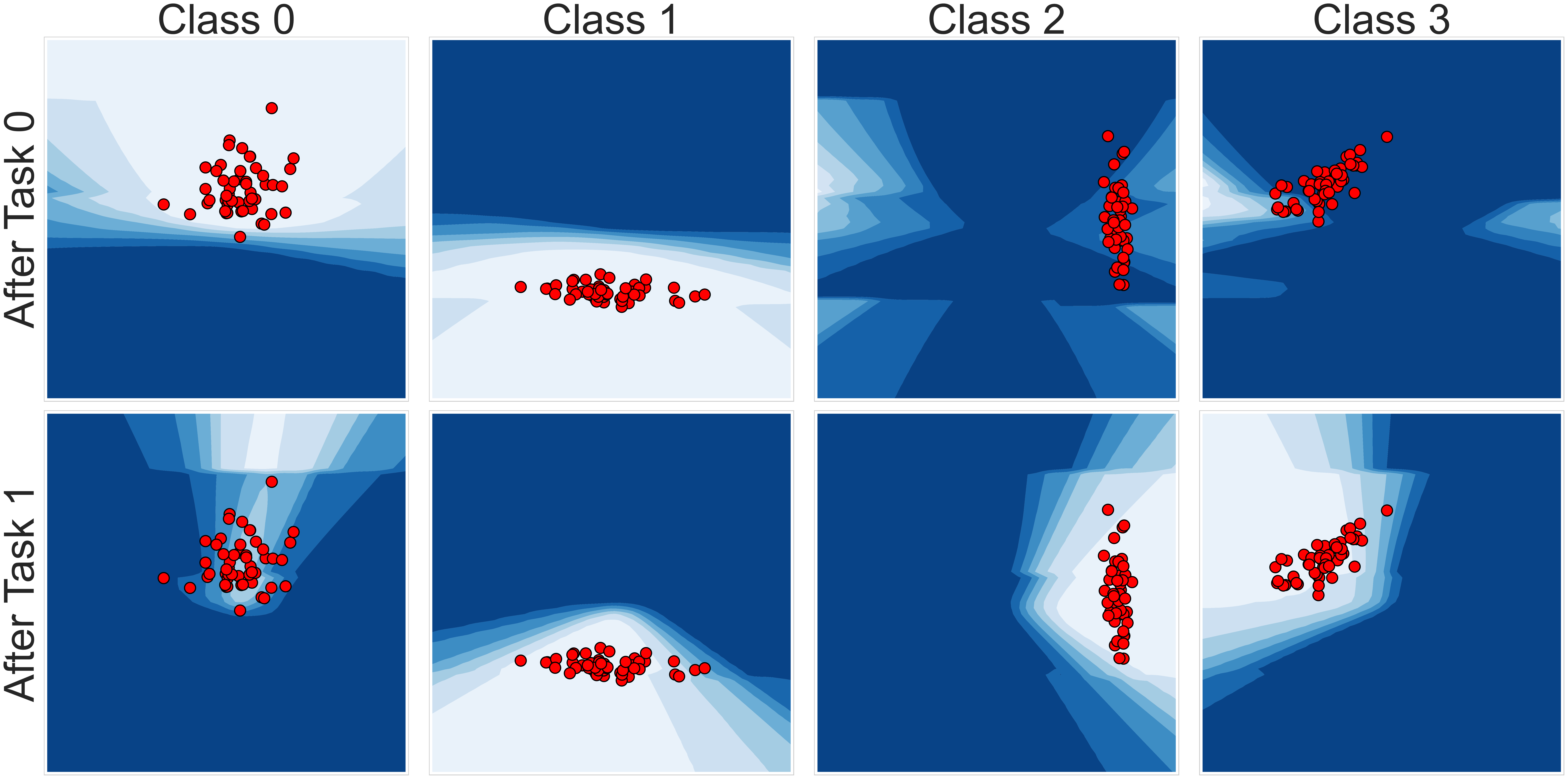} \\
(b) VCL (with coreset size 10)
\end{tabular}
\caption{
By first, only training on classes $0/1$ (\textbf{Task 0}), and next, only 
training on classes $2/3$ (\textbf{Task 1}), we show the posterior predictive 
density surface (brighter is higher) for a four-way classifier on the synthetic 
dataset (marked \tikzcircle[fill=red]{2.5pt}) from \cref{fig:toy_data}. 
(a) VAR-GPs preserve information from Task 0 even after training on Task 1, i.e. 
prevent \emph{catastrophic forgetting} (seen as bright regions around data from 
Task 0). 
(b) VCL is overconfident in its predictions and is structurally less stable 
(seen as large variations in predictive probabilities between Task 0 and Task 1).
}
\label{fig:toy_2d}
\end{figure}

Conventional training methods, however, tend to perform poorly when it comes to 
balancing \emph{rigidity}: the inability to adapt to new experience, and 
\emph{plasticity}: the tendency to forget past experience. For example, neural 
network training is known to be vulnerable to \emph{catastrophic forgetting}, 
causing any data distribution shift to override past learning 
\citep{mccloskey1989catastrophic,ratcliff1990connectionist}. Many approaches 
have been pursued to address these issues, and among them, the Bayesian 
framework is arguably the most general and principled. By allowing us to 
sequentially construct posteriors after observing data online, Bayes' theorem 
provides a coherent framework to build continual learning algorithms, and avoid 
pitfalls like catastrophic forgetting; this principle is seen in 
\citet{sato2001online} for online model selection, and in
\citet{broderick2013streaming} for inference on streaming data. More recently, 
much literature has been devoted to scalable posterior updating schemes with 
neural networks \textemdash ~regularization using a Laplace approximation with 
a diagonal Fisher information matrix \citep{kirkpatrick2017overcoming}, using 
approximate path integral of gradient vector fields \citep{zenke2017continual}, 
using recursive variational posterior approximations 
\citep{nguyen2017variational,swaroop2019improving,ahn2019uncertainty}, 
targeting adaptive capacity in Bayesian neural networks 
\citep{kessler2019hierarchical}, and using episodic memory 
\citep{lopez2017gradient, rebuffi2017icarl}. Alternatively, the effectiveness 
of Gaussian Process (GP) priors is demonstrated by \citet{csato2002sparse} for 
online regression, and by \citet{bui2017streaming} for streaming data. Despite 
favorable properties, GPs remain under-explored for modern continual learning 
tasks.

We take a step in this direction and propose \emph{Variational Auto-Regressive 
Gaussian Processes} (VAR-GPs) for continual learning. By utilizing scalable 
sparse approximations for Gaussian processes \citep{titsias2009variational, 
hensman2013gaussian, hensman2015scalable} and advances in variational inference 
\citep{hoffman2013stochastic}, we make the following key contributions 
\textemdash ~i) a generalized continual variational lower bound for sequential 
datasets with a natural interpretation, ii) a novel auto-regressive variational 
distribution for continual learning with GPs, iii) two fruitful connections to 
results in Bayesian inference, Expectation Propagation (EP) and orthogonal 
inducing point approximation, and iv) evidence on the effectiveness of 
hyper-priors.

The rest of the paper is organized as follows \textemdash~
\cref{sec:relatedwork} puts prior work in context and highlights the key 
similarities or differences. \cref{sec:background} summarizes the relevant 
background needed to develop VAR-GPs. \cref{sec:vargp} derives VAR-GPs for 
continual learning and describes connections to two existing results in 
Bayesian inference. \cref{sec:experiments} develops an intuitive 
characterization of how VAR-GPs avoid catastrophic forgetting, supported by 
empirical results and thorough ablation studies. Finally, \cref{sec:summary} 
concludes with a discussion on limitations and future directions.

\section{Related Work} \label{sec:relatedwork}

Our work aligns with the key desiderata for continual learning outlined by 
\citet{farquhar2018towards} and \citet{van2019three}. In particular, unlike 
prior work \citep{kirkpatrick2017overcoming,zenke2017continual,
nguyen2017variational,swaroop2019improving} where task identity is revealed by 
a multi-head architecture, we tackle a considerably harder and more realistic 
continual learning setup where task identity is unknown and rely on approximate 
Bayesian inference for all learning. While the broader literature focuses on 
many aspects of CL like vanilla transfer learning \citep{li2017learning}, 
adaptive model capacity, and episodic memory, we emphasize on developing an 
automated posterior updating mechanism for continual learning.

Continual learning with neural networks relies on regularization of the 
parameters through the previous approximate posterior (or the prior in absence 
of learning) - \emph{Elastic Weight Consolidation} (EWC) 
\citep{kirkpatrick2017overcoming} uses Laplace approximations with a diagonal 
Fisher Matrix, \emph{Synaptic Intelligence} (SI) \citep{zenke2017continual} 
uses approximate path integral of gradient vector fields, and \emph{RWalk} 
\citep{chaudhry2018riemannian} unifies these methods from a KL-divergence 
perspective. \citet{ritter2018online} use Kronecker-factored approximated 
Laplace approximation for tractability instead. By directly estimating a 
recursive approximate variational posterior, \emph{Improved Variational 
Continual Learning} (VCL) \citep{nguyen2017variational, swaroop2019improving} 
proves to be one of the most competitive methods, and forms our main baseline. 
Although, in principle VCL relies on regularizing only w.r.t the last 
constructed posterior, in practice episodic memory \citep{swaroop2019improving,
shin2017continual,rebuffi2017icarl} of real samples is used to limit 
catastrophic forgetting. We avoid this ad-hoc data subset selection problem by 
naturally incorporating past tasks into our learning objective.

Model capacity has been investigated by \emph{Progressive Neural Networks} 
\citep{rusu2016progressive,schwarz2018progress}, which adapt by rewiring neural 
networks. \citet{kessler2019hierarchical} provide a fully Bayesian treatment 
through Indian Buffet Process priors. While, assessing model capacity required 
for both batch and continual learning remains an open problem, including in 
GP-based models, we emphasize only on an automated inference scheme in this 
work.

\citet{csato2002sparse,Csat2002GaussianPS} are seminal 
contributions towards utilizing Gaussian processes (GPs) in online regression. 
In the spirit of streaming variational Bayes framework 
\citep{broderick2013streaming}, \citet{bui2017streaming} demonstrate the
effectiveness of GPs for a single-task online regression problems. 
\citet{moreno2019continual} adapt recursive construction from VCL 
\citep{nguyen2017variational} to multi-task GPs. To the best of our knowledge, 
no prior work addresses modern continual learning benchmarks and desiderata 
with GPs alone. More recently, FCRL \citep{titsias2019functional} applies 
functional regularization to neural networks using sparse GPs, unifying 
memory-based \citep{lopez2017gradient,rebuffi2017icarl} and Bayesian methods 
through inducing points selection. Unlike FCRL, our proposed approach 
naturally models cross-task covariances through structured 
approximations, and allows joint optimization of both the variational 
approximation and inducing inputs using a unified objective.

\section{Background} \label{sec:background}

\paragraph{Exact Gaussian processes} We assume basic familiarity with Gaussian 
processes (GPs) \citep{rasmussen2003gaussian}. By expressing priors over the 
function space, GPs provide a flexible non-parametric framework to perform 
probabilistic inference. For a dataset $\dset$ of size $N$, we model the 
relationship between a collection of inputs ${\Xvec = \{ \xvec_i \}_{i=1}^N}$
and targets ${\yvec = \{ y_i \}_{i=1}^N}$ using a Gaussian Process prior 
${f \mid \hpvec \sim \gp{\mu_\hpvec(\cdot), k_\hpvec(\cdot, \cdot)}}$ and a 
likelihood model $p(\yvec \mid f(\Xvec))$. The prior is fully defined by a mean 
function $\mu_\hpvec$ and a covariance function $k_\hpvec$, where $\hpvec$ is a 
small set of hyperparameters. Often, we will denote the covariance matrix as 
$\Kvec_{\Xvec, \Zvec}$ to be explicit about the two sets of inputs $\Xvec$ and 
$\Zvec$ that generate the matrix, keeping dependence on $\hpvec$ implicit. As 
typical in literature, we set the mean function to zero.

The intertwined goals of exact inference are to infer the posterior distribution 
${p(f \mid \dset,\hpvec)}$ and obtain the marginal likelihood 
${p(\yvec \mid \Xvec,\hpvec) = \int p(\yvec \mid  f(\Xvec)) p(f \mid \hpvec) 
df}$. The posterior distribution can be used to obtain the predictive 
distribution for a novel input $\xvec_\star$, as ${p(\yvec_\star | \xvec_\star, 
\dset, \hpvec) = \int p(\yvec_\star |  f(\xvec_\star)) p(f | \dset,\hpvec) df}$, 
while the marginal likelihood can be used for model selection. For instance, a 
Gaussian likelihood model with a diagonal covariance ${p(\yvec | f, \Xvec) = 
\mathcal{N}(\yvec; f(\Xvec), \sigma^2_y\mathbf{I})}$ leads to closed-form 
expressions for predictive distribution, which is a Gaussian defined by the 
following mean and variance,
\begin{align}
\mu_\star &= \Kvec_\star^\top (\Kvec + \sigma^2_y\mathbf{I})^{-1}\yvec \label{eq:gp_pred_mean}~, \\
\sigma^2_\star &= \Kvec_{\star\star} - \Kvec_\star^\top(\Kvec + \sigma^2_y\mathbf{I})^{-1}\Kvec_\star~, \label{eq:gp_pred_var}
\end{align}

where ${\Kvec_{\star\star} = k_{\hpvec}(\xvec_\star,\xvec_\star)}$, and
the covariance matrices are given by
${\left(\Kvec_{\star}\right)_i = k_{\hpvec}(\xvec_i,\xvec_\star)}$, and 
${\left(\Kvec\right)_{ij} = k_\hpvec(\xvec_i,\xvec_j)}$. This model can be 
extended for $K$-way classification by building $K$ independent functions and 
using a \emph{softmax} likelihood function such that class probabilities are 
given by ${p(y = k \mid f_1, \dots, f_K, \xvec) \propto \exp\{ f_k(\xvec) \}}$.

Unfortunately, exact inference is analytically and computationally intractable 
for many models and datasets of interest. Even for the closed-form solution 
above for GP regression, the computational complexity is $\bigo{N^3}$ for 
inference \& learning, and $\bigo{N^2}$ for every subsequent prediction. We will 
rely on sparse variational approximations to sidestep both intractabilities.

\paragraph{Sparse Variational Gaussian processes} Among various sparse 
approximations of Gaussian processes \citep{quinonero2005unifying,
bui2017unifying}, our continual learning algorithm builds on inducing-point 
methods \citep{titsias2009variational,hensman2013gaussian, hensman2015scalable}. 
Following \citet{hensman2015scalable}, we introduce $M$ \emph{inducing outputs} 
${\uvec = \{ u_i \}_{i=1}^M}$, and corresponding \emph{inducing inputs} 
${\Zvec = \{ \zvec_i \}_{i=1}^M}$. We view ${\uvec, \Zvec}$ not simply as a
collection of points, but $\uvec$ as values of a continuous function $f$ 
evaluated at $\Zvec$, i.e. $\uvec = f(\Zvec)$. By noting that the underlying 
function can be decomposed as ${f=\{f_{\neq\uvec},\uvec\}}$, the joint 
distribution can be equivalently written as,
\begin{align}
\begin{split}
p(\yvec, f \mid \Xvec, \hpvec) =&~  p(\yvec \mid f, \Xvec) \\
&~ p(f_{\neq \uvec} \mid\uvec, \hpvec ) p(\uvec \mid \Zvec, \hpvec)~.
\end{split} \label{eq:svgp_joint}
\end{align}

The structured variational approximation is judiciously chosen to be 
${q(f) = p(f_{\neq \uvec} |\uvec, \hpvec) q(\uvec)}$. This leads to a 
cancellation of difficult terms, yielding the following variational lower bound, 
\begin{align}
\mathcal{F}(q, \hpvec) =&~ \mathbb{E}_{q(f)}\left[ \log{\frac{p(\yvec \mid f) \cancel{p\left(f_{\neq \uvec} \mid \uvec, \hpvec \right)} p(\uvec \mid \Zvec, \hpvec) }{\cancel{p(f_{\neq \uvec} \mid \uvec, \hpvec)}q(\uvec)}} \right] \nonumber \\
\begin{split}
\mathcal{F}(q, \hpvec) =&~ \sum_{i=1}^N \mathbb{E}_{q(f)} \left[ \log{p\left(y_i \mid f, \xvec_i \right)} \right] \\
&~ - \kl\left[ q(\uvec) \mid\mid p(\uvec \mid \hpvec) \right]~.	
\end{split}
\end{align}

The first term in this bound remains intractable for a general likelihood and 
large $N$. However, it can be approximated by simple Monte Carlo with 
reparameterisation gradients and data subsampling \citep{jordan1999introduction,
hoffman2013stochastic, hensman2013gaussian}. \citet{bui2017streaming} extend 
this bound for streaming data, and show good performance on time series 
regression. This approach, however, employs a single set of inducing points 
which is \emph{shared} across multiple time steps or tasks and, as a result, the 
inducing points are either too rigid to adapt, or too quickly moving to regions 
of new data. We demonstrate this pitfall in \cref{sec:experiments}. Equipped 
with this background, we are now ready to derive the proposed approximation, 
VAR-GPs. 

\section{Variational Auto-Regressive Gaussian Processes} \label{sec:vargp}

Consider datasets ${\{\dset^{(1)}, \dots, \dset^{(T)}\}}$, of sizes 
${\{N_1, \dots, N_T\}}$ respectively, for $T$ different but related tasks. These 
tasks are observed sequentially and only once. We want a model which 
performs well not only on the current task, but also sustains performance on 
previous tasks. In subsequent notations, we identify task-specific quantities 
with corresponding task numbers, $\{1, \dots, T\}$.

\subsection{Learning the First Task} \label{sec:first_vfe}

We first extend the model in \cref{eq:svgp_joint} to incorporate a prior over 
the hyperparameters $p(\hpvec)$ and choose the corresponding variational 
posterior ${q_1(f, \hpvec) = p\left(f_{\neq \uvec_1} \mid \uvec_1, \hpvec 
\right)q(\uvec_1)q_1(\hpvec)}$. While the correlation between $\uvec$ 
and $\hpvec$ is ignored in the approximate posterior, the dependencies between 
$\hpvec$ and the remaining function values are retained through the conditional 
prior term. Learning the first task using $\mathcal{D}^{(1)}$ is thus a direct 
extension of the sparse variational approach in \cref{sec:background}, leading 
to the following bound,
\begin{align}
\begin{split}
\mathcal{F}(q_1) =& \sum_{i=1}^{N_1} \mathbb{E}_{q_1(f,\hpvec)} \left[\log{p\left(y_i^{(1)} \mid f, \xvec_i^{(1)} \right)} \right] \\
&- \kl\left[ q_1(\hpvec) \mid\mid p(\hpvec) \right] \\
&- \mathbb{E}_{q_1(\hpvec)} \left[ \kl\left[ q(\uvec_1) \mid\mid p(\uvec_1 \mid \hpvec) \right] \right]	~.
\end{split}\label{eq:elbo_1}
\end{align}

We stress that quantifying uncertainty in hyperparameters through $q_1(\hpvec)$ 
is important for continual learning, as supported by our ablation studies in 
\cref{sec:ablations}.

\subsection{Generalized Continual Variational Lower Bound}  
\label{sec:general_cont_vfe}

Instead of reusing and updating the same set of inducing inputs and outputs for 
all subsequent tasks as in \citet{bui2017streaming}, we introduce a separate set 
of inducing inputs and outputs, ${\{\Zvec_t, \uvec_t\}}$, for each task.
Consider the $t$-th task and $t>1$, using notation in a spirit similar to 
\cref{eq:svgp_joint}, the function can be split into, ${f = \{f_{\neq 
\uvec_{\leq t}}, \uvec_{< t}, \uvec_t \}}$ that describes a decomposition in 
terms of all past inducing outputs $\uvec_{<t}$ and current inducing outputs 
$\uvec_t$. This leads to an approximate running joint for $\dset^{(t)}$ 
conditioned on past data as,
\begin{align}
\begin{split}
p(\yvec^{(t)}, f, \hpvec \mid \Xvec^{(t)}, \mathcal{D}^{(< t)}) \approx & \prod_{i=1}^{N_t} p(y_i^{(t)} \mid f, \xvec_i^{(t)}) \\
& p(f_{\neq \uvec_{\leq t}} \mid \uvec_{\leq t}, \hpvec) \\
& p(\uvec_t \mid \Zvec_t, \uvec_{< t}, \hpvec) \\
& q(\uvec_{< t} \mid \Zvec_{< t}, \hpvec) \\
& q_{t-1}(\hpvec) ~.
\end{split}\label{eq:general_cont_model}
\end{align}

We mirror the form of the prior in the approximate posterior, in a similar 
fashion to the first task, to obtain the following variational distribution, 
\begin{align}
\begin{split}
q_t(f, \hpvec) =&~p(f_{\neq \uvec_{\leq t}} \mid \uvec_{\leq t}, \hpvec) q(\uvec_t \mid \Zvec_t, \uvec_{< t}, \Zvec_{< t}, \hpvec) \\
&~q(\uvec_{< t} \mid \Zvec_{< t}, \hpvec) q_{t}(\hpvec)~.	
\end{split} \label{eq:q_general}
\end{align}

This structured form leads to the cancellation of ${p(f_{\neq \uvec_{\leq t}} 
\mid\uvec_{\leq t}, \hpvec)}$ and ${q(\uvec_{< t} \mid \Zvec_{< t}, \hpvec)}$ 
to arrive at the generalized continual variational lower bound,
\begin{align}
\begin{split}
\mathcal{F}(q_t) =& \sum_{i=1}^{N_t} \mathbb{E}_{q_t(f,\hpvec)}\left[ \log{p(y_i^{(t)} \mid f, \xvec_i^{(t)})} \right] \\
&- \kl\left[ q_t(\hpvec) \mid\mid q_{t-1}(\hpvec) \right] \\
&- \mathbb{E}_{q_t(\hpvec)q(\uvec_{< t} \mid \Zvec_{< t}, \hpvec)}\left[ \mathfrak{D}_t \right] ~,
\end{split} \label{eq:elbo_general}
\end{align}
where,
\begin{small}\begin{align*}
\mathfrak{D}_t = \kl \left[q(\uvec_t \mid \Zvec_t, \uvec_{< t}, \Zvec_{< t}, \hpvec) \mid\mid p(\uvec_t \mid \Zvec_t, \uvec_{< t}, \Zvec_{< t}, \hpvec) \right]~.
\end{align*}\end{small}We emphasize that any dependence on $\uvec_t$ is 
accompanied by corresponding $\Zvec_t$ as presented in \cref{eq:gp_pred_mean} 
and \cref{eq:gp_pred_var}, but may often keep it implicit for concise notation.

As our learning objective, the maximization of \cref{eq:elbo_general} takes a 
natural interpretation \textemdash~we maximize the likelihood of current data 
$\mathcal{D}^{(t)}$, subject to a $\kl$-regularization that balances past 
posterior and new data. The regularization term involves hyperparameters and 
current inducing outputs. In practice, tempering the hyperparameter distribution 
may prove helpful under misspecified models \cite{wenzel2020good,
wilson2020bayesian}, which is equivalent to scaling 
${\kl\left[ q_t(\hpvec) \mid\mid q_{t-1}(\hpvec) \right]}$ by a positive scalar 
$\beta$. The new inducing points ${\{\Zvec_t, \uvec_t \}}$ are used to explain 
new parts of the data space while the old ones ${\{ \Zvec_{<t}, \uvec_{<t}\}}$
aim to preserve past experience, as demonstrated in \cref{fig:toy_2d}. Next, we 
develop this intuition further, and experimentally demonstrate in 
\cref{sec:qualitative_analysis}. 

\subsection{Distributional Choices} \label{sec:var_params}

We now detail the parametrizations for all aforementioned distributions. All
prior-related densities, i.e. ${p(\uvec_1 \mid \hpvec)}$, 
${p(\uvec_t \mid \uvec_{< t}, \hpvec)}$, and ${p(f_{\uvec_{\leq t}} \mid 
\Xvec^{(t)}, \uvec_{\leq t}, \hpvec)}$ can be computed by invoking the GP prior. 
All experiments use the Exponentiated Quadratic kernel such that $\hpvec$ 
includes the $\log$-ARD lengthscales and a $\log$-scale factor (see 
\cref{sec:sq_exp_kernel}). The prior over the $\log$-hyperparameters for the 
first task $p(\hpvec)$ is assumed to be a standard Normal, 
${\mathcal{N}(\hpvec; \mathbf{0}, \mathbf{I})}$.

The variational distribution of the $\log$-hyperparameters is assumed to be a 
mean-field Gaussian, i.e. parametrized by a diagonal covariance, 
${q_t(\hpvec) = \mathcal{N}(\hpvec; \bm{\mu}_t, \texttt{diag}(\bm{\sigma_t}))}$. 
These choices allow for closed-form $\kl$ computations in 
\cref{eq:elbo_1,eq:elbo_general}. The variational distribution over inducing 
outputs for the first task is parametrized as 
${q(\uvec_1) = \mathcal{N}(\uvec_1; \mvec_1, \Svec_1)}$ for a set of $M_1$
inducing outputs, as standard in sparse variational GPs using mean $\mvec_1$ and
covariance matrix $\Svec_1$.

One of our key contributions is an auto-regressive parametrization for the 
variational distribution given by ${q(\uvec_{\leq t} \mid \hpvec) = 
q(\uvec_1) \prod_{j=2}^{t} q(\uvec_j \mid \uvec_{< j}, \hpvec)}$, where 
${q(\uvec_t \mid \uvec_{< t}, \hpvec) = 
\mathcal{N}(\uvec_t; \Kvec_{\Zvec_t, \Zvec_{< t}} \Kvec_{\Zvec_{< t}, 
\Zvec_{< t}}^{-1} \uvec_{< t} + \mvec_t, \Svec_t)}$ relies on $M_t$ inducing 
variables for all subsequent tasks $t>1$. As a consequence of this structure, 
$\mathfrak{D}_t$ becomes independent of $\uvec_{< t}$ and avoids sampling 
variance introduced by samples from $q(\uvec_{< t}\mid \Zvec_{<t},\hpvec)$. A 
simplifying trick to compute ${q(\uvec_{< t} \mid \hpvec)}$ using conditional 
Gaussian identities (see \cref{sec:auto_regressive_trick}). We further note that 
(i) the marginal density for inducing outputs $\uvec_t$ when $t>1$ is 
non-Gaussian, and (ii) even though ${\Zvec_t, \mvec_t, \Svec_t}$ are kept fixed 
after training on the $t$-th task, the marginal density $q(\uvec_t)$ can still 
change over time due to the change in $q(\hpvec)$.

Next, \cref{sec:connections} provide two fruitful connections of the 
proposed structured variational approximation to existing literature.

\subsection{Connections to Expectation Propagation and Orthogonal Inducing Points}
\label{sec:connections}

\paragraph{Structured EP Factor Approximation leads to an Auto-regressive Approximate Posterior} 
For the approximate running joint in \cref{eq:general_cont_model}, we can 
introduce an approximation to the posterior as,
\begin{align}
\begin{split}
q_t(f, \hpvec) \propto & \left[\prod_{i=1}^{N_t} \mathbf{g}_i^{(t)}(\hpvec) \mathbf{h}_i^{(t)}(\uvec_t)\right] p(f_{\neq \uvec_{\leq t}} \mid \uvec_{\leq t}, \hpvec) \\
&~ q(\uvec_{< t} \mid \hpvec) p(\uvec_t \mid \uvec_{< t}, \hpvec) q_{t-1}(\hpvec),	
\end{split}
 \label{eq:ep_factor}
\end{align}

where the difficult likelihood term, ${\prod_{i=1}^{N_t} p(y_i^{(t)} \mid f, 
\xvec_i^{(t)})}$, is approximated by ${\prod_{i=1}^{N_t}  \mathbf{g}_i^{(t)}
(\hpvec) \mathbf{h}_i^{(t)}(\uvec_t)}$ such that $\mathbf{g}_i$ and 
$\mathbf{h}_i$ are the approximate contributions of each likelihood term to the 
posterior. Expectation Propagation (EP) \citep{DBLP:conf/uai/Minka01} then
proceeds by repeating the following steps to convergence: i) remove the 
approximate contributions $\mathbf{g}_i$ and $\mathbf{h}_i$ from the posterior 
to form the cavity for $i$-th datum, ii) merge the cavity with 
${p(y_i^{(t)} \mid f, \xvec_i^{(t)})}$ to form the tilted distribution 
${\tilde{p}_i}$, iii) minimize the divergence ${\kl[\tilde{p}_i \mid\mid q_t]}$ 
to find a new approximate posterior, and iv) obtain the new approximate factors 
$\mathbf{g}_i$ and $\mathbf{h}_i$ by removing the cavity from the new posterior. 

We are, however, not interested in running EP, but only in the form of the 
approximate posterior induced by EP. Merging relevant terms in the approximate 
posterior as
\begin{align}
\begin{split}
q_t(f, \hpvec) \propto &~ p(f_{\neq \uvec_{\leq t}} \mid \uvec_{\leq t}, \hpvec) \\
&~ q(\uvec_t \mid \uvec_{< t}, \hpvec)q(\uvec_{< t} \mid \hpvec) \\
&~ q_{t}(\hpvec)~,
\end{split}	
\end{align}
where,
\begin{small}
\begin{align*}
q_{t}(\hpvec) &\propto {q_{t-1}(\hpvec)}\prod_{i=1}^{N_t} {\mathbf{g}_i^{(t)}(\hpvec)}~, \\
{q(\uvec_t \mid \uvec_{< t}, \hpvec)} &\propto p(\uvec_t \mid \uvec_{< t}, \hpvec) \prod_{i=1}^{N_t} \mathbf{h}_i^{(t)}(\uvec_t)~.
\end{align*}	
\end{small}

We find that the auto-regressive factor is given by 
${p(\uvec_{t} \mid \uvec_{< t}, \hpvec) = 
\mathcal{N}(\uvec_t; \mathbf{A}_t \uvec_{<t}, \mathbf{C}_{t})}$, where we define
${\Avec_t \defeq 
\Kvec_{\Zvec_t, \Zvec_{< t}} \Kvec_{\Zvec_{< t}, \Zvec_{< t}}^{-1}}$, and
${\mbf{C}_t \defeq \Kvec_{\Zvec_t, \Zvec_t} - \Kvec_{\Zvec_t, \Zvec_{< t}} 
\Kvec_{\Zvec_{< t}, \Zvec_{< t}}^{-1} \Kvec_{\Zvec_{< t}, \Zvec_t}}$.

Now, consider a Gaussian factor approximation ${\mathbf{H}_t(\uvec_t) = 
\prod_{i=1}^{N_t} \mathbf{h}_t(\uvec_t) = \mathcal{N}(\uvec_t; \mu, \Sigma)}$.
By multiplying ${\mathbf{H}_t(\uvec_t)}$ with
${p(\uvec_{t} \mid \uvec_{< t}, \hpvec)}$ and renormalizing, we arrive at 
${q(\uvec_t \mid \uvec_{<t}, \hpvec) = 
\mathcal{N}(\uvec_t; \tilde{\mu}, \tilde{\Sigma})}$, 
where ${\widetilde{\Sigma}^{-1} = \Sigma^{-1} + \mathbf{C}_{t}^{-1}}$ and 
${\widetilde{\Sigma}^{-1} \tilde{\mu} = 
\Sigma^{-1} \mu + \mathbf{C}_{t}^{-1} \mathbf{A}_t \uvec_{<t}}$. The conditional 
posterior mean can be reformulated as ${\widetilde{\mu} = 
\mathbf{A}_t \uvec_{< t} + (\mathrm{I} + 
\Sigma\mathbf{C}_t^{-1})^{-1} (\mu - \mathbf{A}_t \uvec_{< t})}$, so that we can 
parameterize the conditional posterior instead of the factor
$\mathbf{H}_t(\uvec_t)$, giving us ${q(\uvec_t \mid \uvec_{< t}, \theta) =
\mathcal{N}(\uvec_t; \mathbf{A}_t \uvec_{< t} + \mvec, \Svec)}$. This is 
equivalent to our auto-regressive parametrization.

\paragraph{Equivalence to Orthogonal Inducing Points}
Our auto-regressive parametrization is also exactly equivalent to the 
\emph{orthogonal inducing points} formulation when $T=2$ 
\citep[\S{3.3}]{shi2020sparse}. A variational approximation over two set of 
orthogonal inducing points, $\uvec$ and $\vvec$, is presented as
\begin{align}
q(\uvec, \vvec) = \mathcal{N}\left(\begin{bmatrix} \uvec \\ \vvec \end{bmatrix}; \begin{bmatrix} \mvec_\uvec \\ \Kvec_{\vvec\uvec} \Kvec_{\uvec\uvec}^{-1} \mvec_\uvec \end{bmatrix}, \Svec \right)~,
\end{align}
where,
\begin{small}
\begin{align*}
\mathbf{\Sigma} \defeq \begin{bmatrix} \Svec_\uvec & \Svec_\uvec \Kvec_{\uvec\uvec}^{-1} \Kvec_{\uvec\vvec} \\ \Kvec_{\vvec\uvec} \Kvec_{\vvec\uvec}^{-1} \Svec_\uvec & \Svec_\vvec + \Kvec_{\vvec\uvec} \Kvec_{\uvec\uvec}^{-1} \Svec_\uvec  \Kvec_{\uvec\uvec}^{-1} \Kvec_{\uvec\vvec}  \end{bmatrix}~.	
\end{align*}	
\end{small}

By appealing to conditional Gaussian identities, the joint can be factored as
${q(\uvec, \vvec) =  q(\uvec) q(\vvec \mid \uvec)}$, such that 
${q(\uvec) = \mathcal{N}(\uvec; \mvec_{\uvec}, \Svec_{\uvec})}$, and 
${q(\vvec \mid \uvec) = \mathcal{N} (\vvec;  
\Kvec_{\vvec\uvec} \Kvec_{\uvec\uvec}^{-1} \uvec + \mvec_\vvec, \Svec_\vvec)}$. 
This conditional is exactly equivalent to our proposal when there are two tasks, 
i.e. ${\uvec \Leftrightarrow \uvec_1}$ and ${\vvec \Leftrightarrow \uvec_2}$.
Intuitively, the second set of inducing variables $\uvec_2$ attempt to explain 
the data space that is not well-explained by the first set $\uvec_1$.
\citet{shi2020sparse} briefly discussed using more than two sets of inducing 
points for a fixed dataset but did not investigate further due to implementation
complexity and the potential small gain beyond two sets. Unlike the batch 
setting of \citet{shi2020sparse}, our work extends the idea of using many sets 
of such inducing points to the continual learning setting.

\section{Experiments} \label{sec:experiments}

Through our experiments, we highlight the qualitative characteristics of the 
derived generalized learning objective \cref{eq:elbo_general}, and provide 
evidence for the competitiveness of VAR-GPs compared to our main baseline 
Improved Variational Continual Learning (VCL) \citep{nguyen2017variational,
swaroop2019improving}, among others. A thorough ablation study demonstrates the 
efficacy of our modeling choices. The full reference implementation of VAR-GPs 
in PyTorch \citep{NEURIPS2019_9015} in publicly available at 
\codeurl.

\begin{figure}[ht]
\centering
\includegraphics[width=0.6\linewidth]{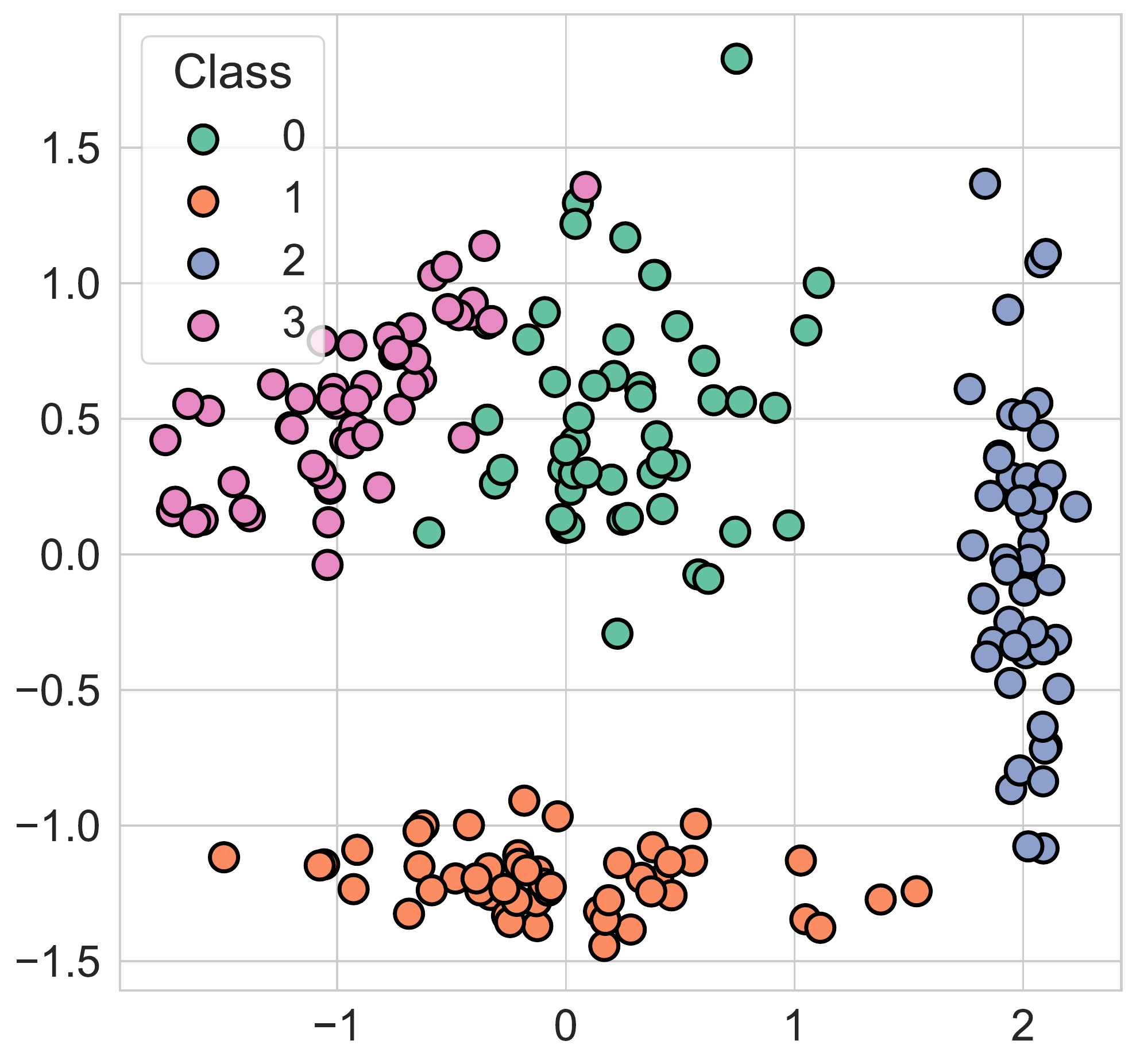}
\caption{A synthetic toy dataset used for the four-way classification problem in \cref{fig:toy_2d}.}
\label{fig:toy_data}
\end{figure}

To mimic a real continual learning setting, the model only observes 
$\dset^{(t)}$ for training and is tested on all the tasks seen so far using 
$\dset^{(\leq t)}$. We track validation accuracy on a subset of the training set
for early stopping. To optimize both variational objectives
\cref{eq:elbo_1,eq:elbo_general}, we use a mini-batch size of $512$ and $3$
samples from variational distribution with the Yogi optimizer 
\citep{zaheer2018adaptive}. Predictive distributions are estimated via $10$ 
Monte-Carlo samples. The training for the $t$-th task is summarized in 
\cref{alg:vargp_train}.

\begin{algorithm}[!ht]
   \caption{VAR-GP per-task training}
   \label{alg:vargp_train}
\begin{algorithmic}
   \STATE \textbf{Input}: Learning rate $\eta$, Batch size $B$, Number of inducing points $M$, Maximum epochs $E$, Task dataset $\mathcal{D}^{(t)}$, hyperparameters: KL tempering factor $\beta$, Early stopping patience of $K$ epochs and tolerance $\delta$
   \STATE \textbf{Output}: Per-task variational approximations for kernel hyper-parameters $\hpvec$ and inducing outputs $\uvec_t$, and inducing inputs $\Zvec_t$
   \STATE Initialize $\Zvec_t \in \mathbb{R}^{M \times D} \subset \Xvec^{(t)} \in \mathbb{R}^{N_t \times D}$
   \FOR{$e$ \textbf{in} $1 \dots E$}
   \FOR{$\{ \xvec_i, y_i \}_{i=1}^B \subset \mathcal{D}^{(t)}$}
   \STATE Compute $\mathcal{F}(q_t)$ \textemdash~ using \cref{eq:elbo_1} for the first task, and \cref{eq:elbo_general} for all subsequent ones.
   \STATE Update $\hpvec$, $\mvec_t$, $\Svec_t$ and $\Zvec_t$ with learning rate $\eta$ and tempering factor $\beta$
   \ENDFOR
   \STATE Compute validation accuracy $A_e$
   \IF{$e > K$ \textbf{and} $|A_e - A_{e-K}| < \delta$}
   \STATE \texttt{break}
   \ENDIF
   \ENDFOR
\end{algorithmic}
\end{algorithm}

All covariance matrices in the variational distributions are modeled as a lower 
triangular Cholesky factors, described in \cref{sec:parametrizing_cov}, 
initialized at an identity matrix for numerical stability, with diagonals 
constrained to be positive during optimization using a \texttt{softplus} 
transform. Further, as described in \cref{sec:sq_exp_kernel}, the kernel 
parameters $\hpvec$, including lengthscales and scale factors modeled as 
\texttt{log}-transforms to maintain positivity. \cref{sec:hypers} lists the 
search space for all hyper-parameters. Results report the mean and one standard 
deviation of five independent trials. Next, we describe the datasets used for 
experiments.

\textbf{Synthetic Classification Dataset} Visualized in \cref{fig:toy_data}, 
we use a synthetic 2-D dataset with four classes in the range 
${x,y \in [-3., 3.]}$ for qualitative assessments. We observe classes in pairs
$0/1$ (Task 0) and $2/3$ (Task 1), each only once. We do not use any tempering 
for this dataset, i.e. ${\beta=1}$.

\textbf{Split MNIST} Following \citet{zenke2017continual}, we consider the full 
$10$-way classification task at each time step but receive a dataset 
$\dset^{(t)}$ of only a subset of MNIST digits in the sequence $0/1$, $2/3$, 
$4/5$, $6/7$, and $8/9$. $10000$ training samples are cumulatively set aside for 
validation set across all tasks. We allocate $60$ inducing points for each task, 
with a learning rate of $0.003$, and $\beta = 10.0$. We remind the reader that 
unlike prior work which uses a multi-head model with task information, we only 
use a single-head to report the classification test accuracy, making the 
benchmark considerably harder and more representative of reality.

\textbf{Permuted MNIST} In this benchmark, we receive a dataset $\dset^{(t)}$ of
MNIST digits at each time step $t$, such that the pixels undergo an unknown but 
fixed permutation. $10000$ samples are set aside for validation. We allocate 
$100$ inducing points for each task, with a learning rate of $0.0037$, and 
$\beta = 1.64$. The first task is fixed to be the unpermuted MNIST to provide an 
upper bound on the performance of subsequent tasks. While prior work 
\citep{zenke2017continual,kirkpatrick2017overcoming} uses this benchmark as an 
indicator of representational capacity of neural networks, we use this to test 
performance under distributional shift.

\subsection{Qualitative Analysis} \label{sec:qualitative_analysis}

Before presenting benchmark results, we seek to intuitively understand how 
VAR-GPs learn and avoid \emph{catastrophic forgetting}. By modeling the 
cross-task covariances through $\mathfrak{D}_t$ in \cref{eq:elbo_general}, we 
are able to bias the learning such that under the given budget of inducing 
points, VAR-GPs retain information about the previous tasks. This is exemplified 
in \cref{fig:toy_2d}. Observing predictive density plots for Classes 0 and 1 
after ``Task 1", we notice that the high density regions from ``Task 0" 
are preserved by VAR-GPs, whereas VCL suffers from large variations in the 
predictions.

\begin{figure}[!ht]
\centering
\begin{tabular}{c}
\includegraphics[width=.95\linewidth]{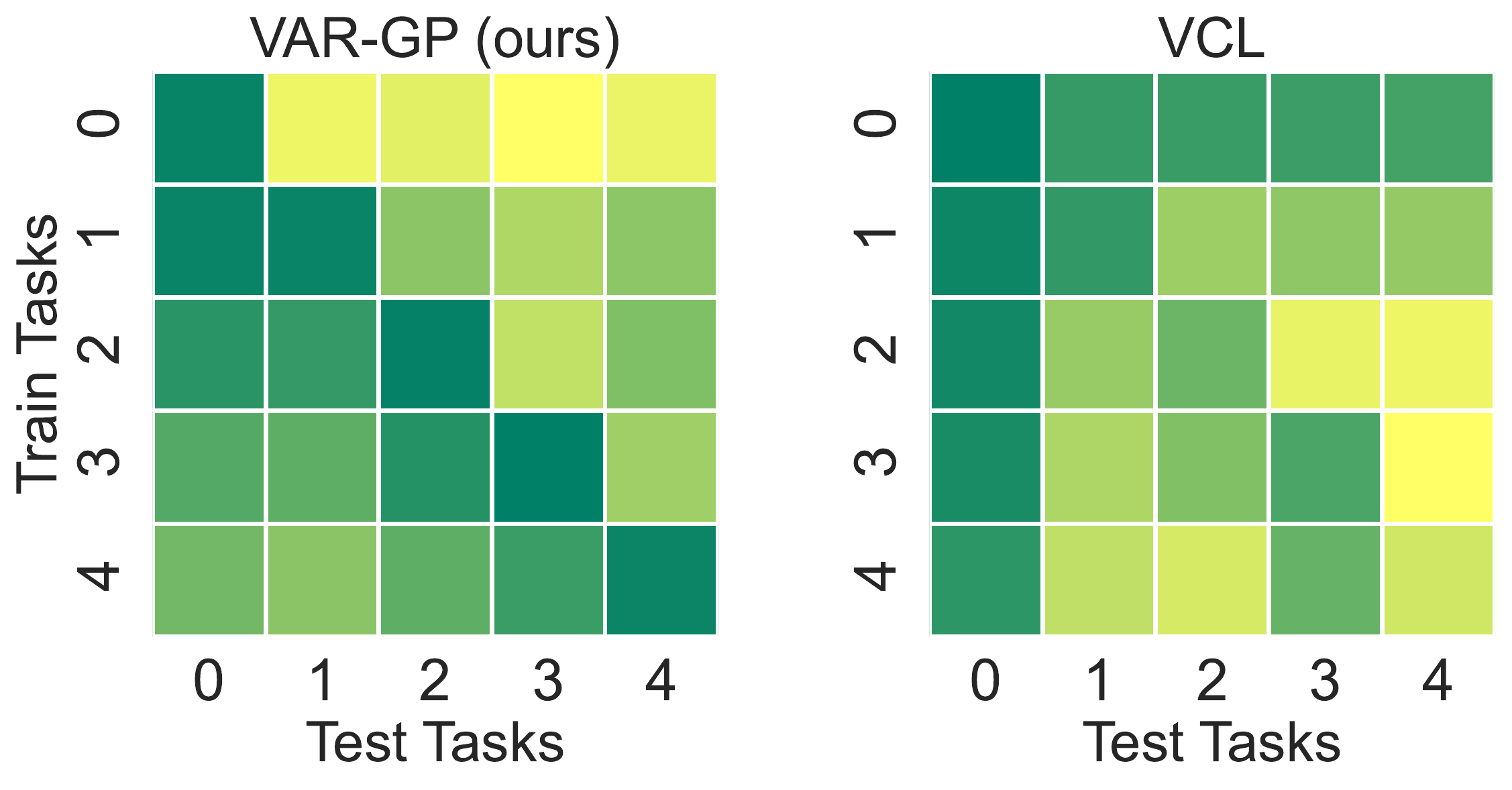}\\
(a) Split MNIST \\ \\
\includegraphics[width=.95\linewidth]{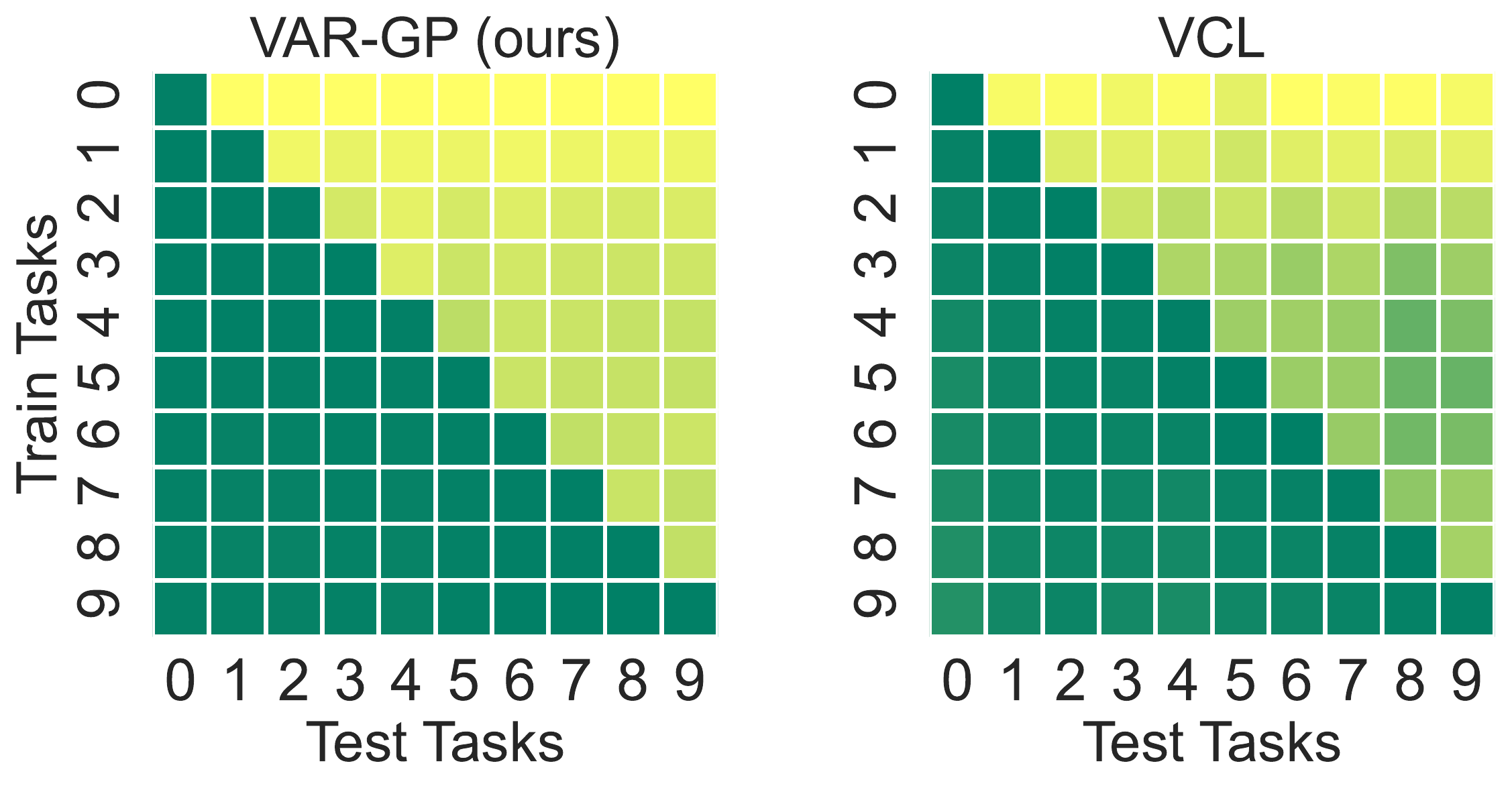} \\
(b) Permuted MNIST
\end{tabular}
\caption{
As we continually train on tasks (shown along the y-axis), we evaluate the mean 
predictive entropy on the test set for all tasks (shown along the x-axis). In 
other words, the upper triangular region shows tasks which are not yet seen 
during training. The values are normalized by the entropy of a random ten-way 
classifier, $\log{10}$. Brighter regions correspond to a higher entropy, i.e.
larger predictive uncertainty. (\textbf{Top}) For Split MNIST, we observe that
VAR-GPs lead to lower forgetting rates through lower predictive entropy in the 
lower triangular regions, and higher uncertainties in upper triangular regions. 
VCL \citep{swaroop2019improving}, on the other hand fails to improve much on
subsequent tasks. (\textbf{Bottom}) For Permuted MNIST, we see that while both 
VAR-GPs and VCL maintain reasonably low catastrophic forgetting on the tasks seen
so far, VCL tends to be overconfident on unseen tasks, i.e. low predictive 
entropy in the upper triangular region.
}
\label{fig:split_perm_mnist_entropies}
\end{figure}

This characteristic is not restricted to the toy dataset. Using mean predictive 
entropy estimates to quantify uncertainty over the test set, we visualize the 
degree of forgetting for benchmark datasets in 
\cref{fig:split_perm_mnist_entropies}. Even as tasks progress, VAR-GPs 
demonstrate information preservation from old tasks by keeping entropy low. In 
contrast, often being overconfident in predictions, VCL keeps entropy low only 
for the old tasks at the expense of new ones.

Finally, an inspection of the optimized inducing points reveals that VAR-GPs 
tend to focus on covering the regions of the current task, as one would 
naturally expect. \cref{fig:viz_smnist_ind_pts} showcases this behavior for 
Split MNIST. We anticipate such space covering behavior from our learning 
objective \cref{eq:elbo_general} due to the cross-correlations modeled in the 
regularization term $\mathfrak{D}_t$.

\bgroup
\setlength\tabcolsep{1pt}
\begin{figure}[!ht]
\centering
\begin{tabular}{c|c|c|c|c}
\includegraphics[width=.19\linewidth]{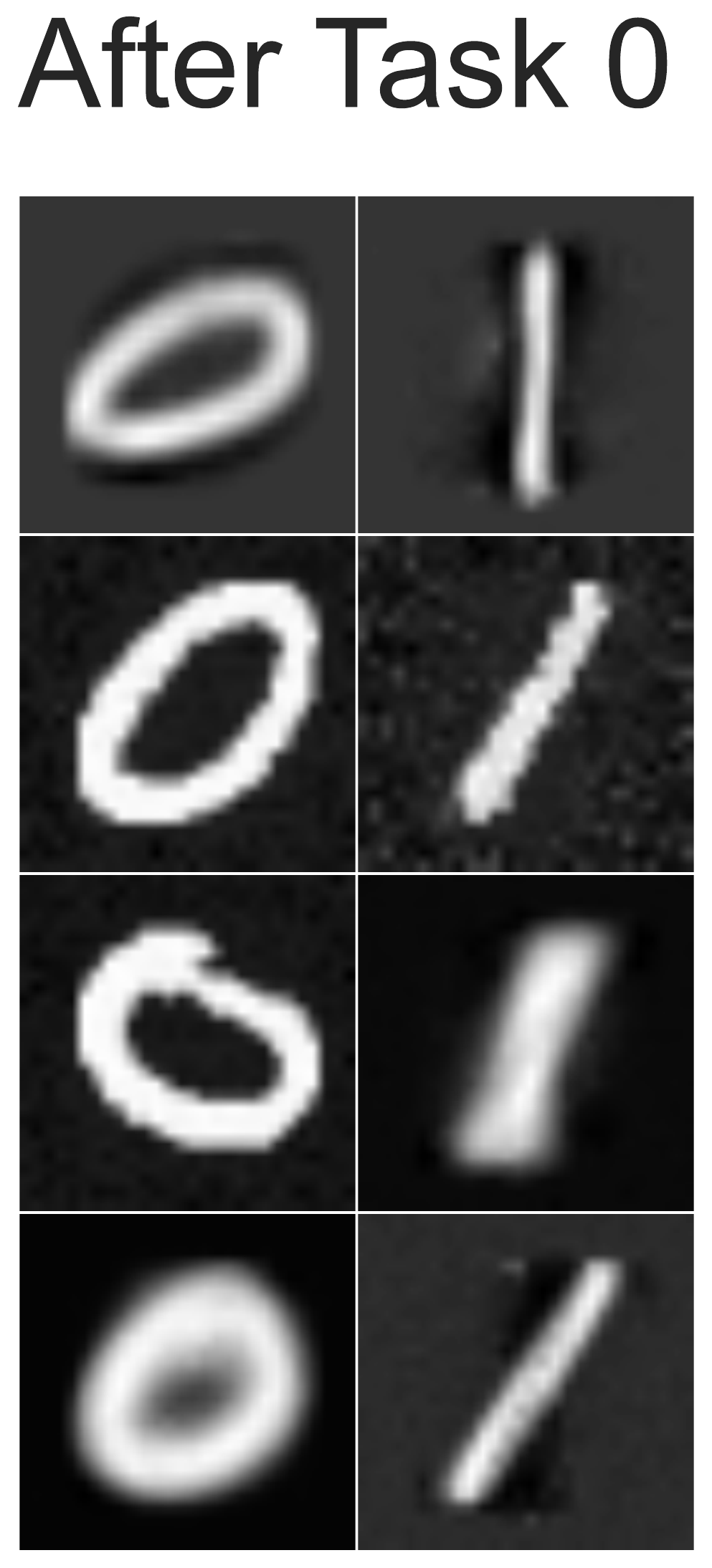} &
\includegraphics[width=.19\linewidth]{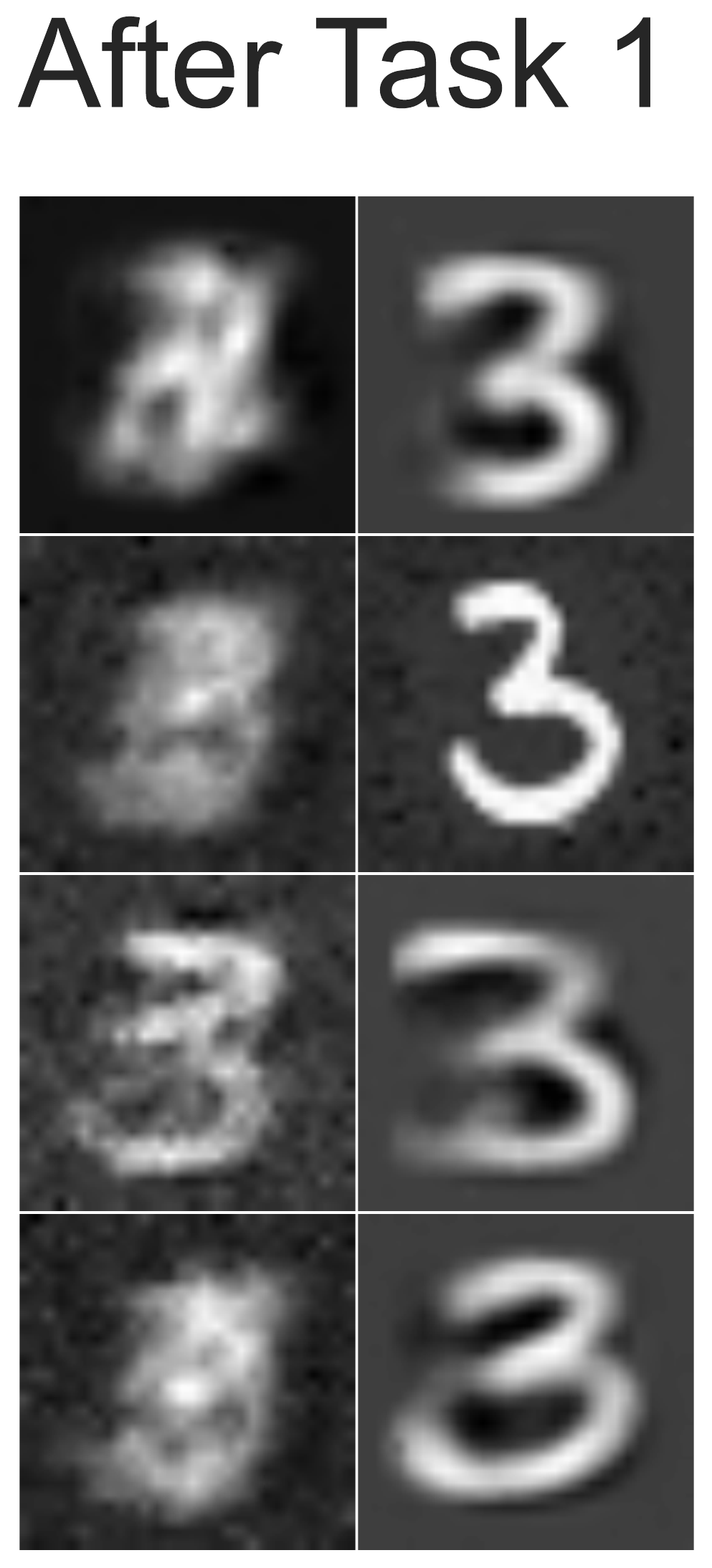} & 
\includegraphics[width=.19\linewidth]{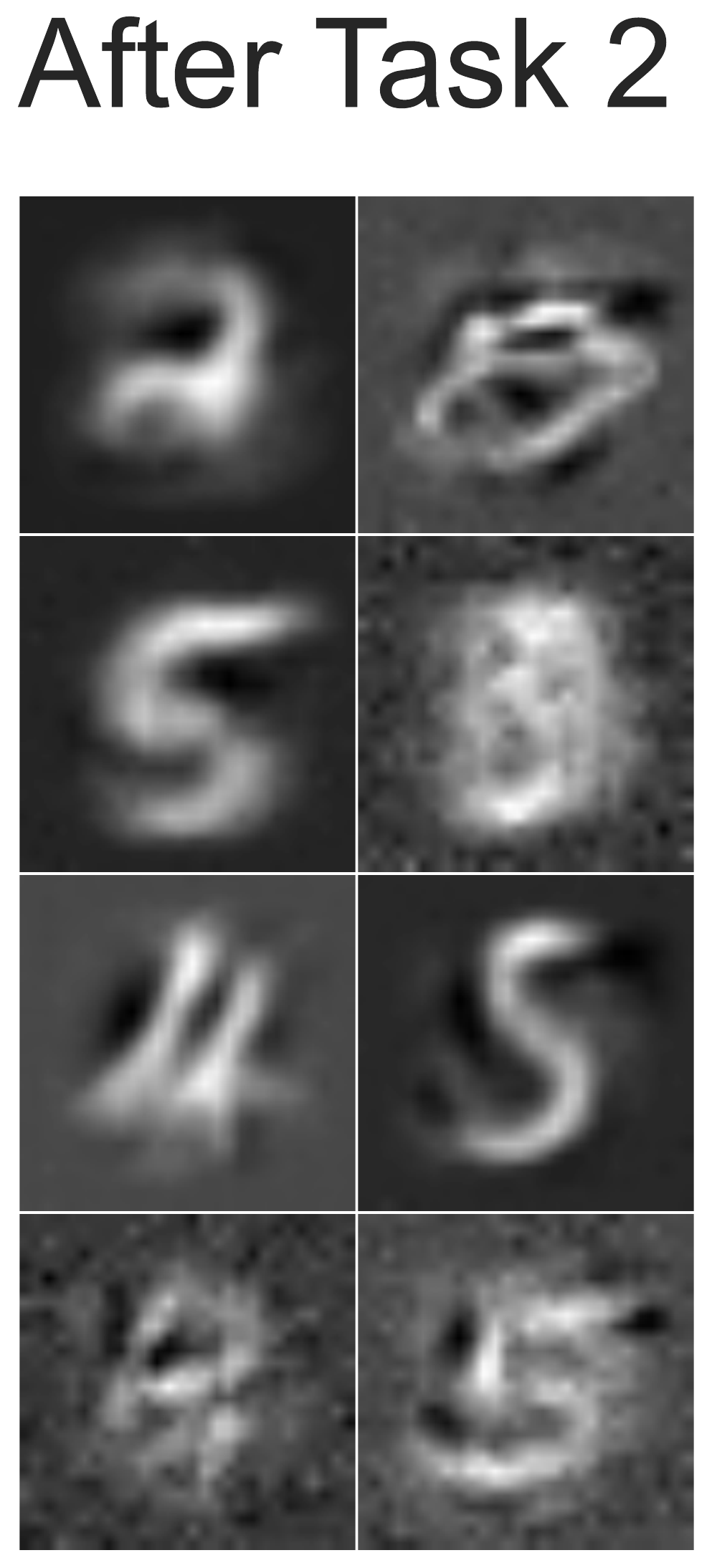} &
\includegraphics[width=.19\linewidth]{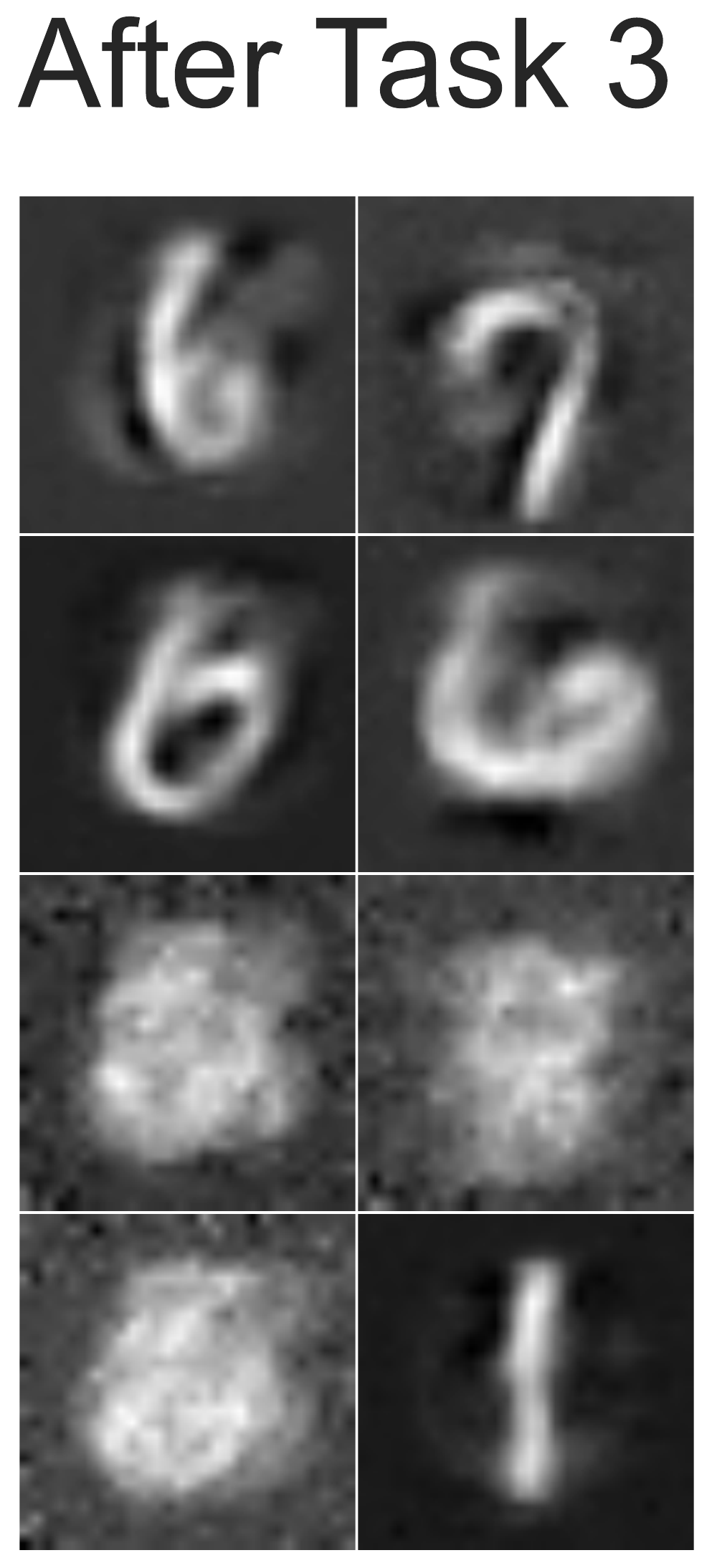} &
\includegraphics[width=.19\linewidth]{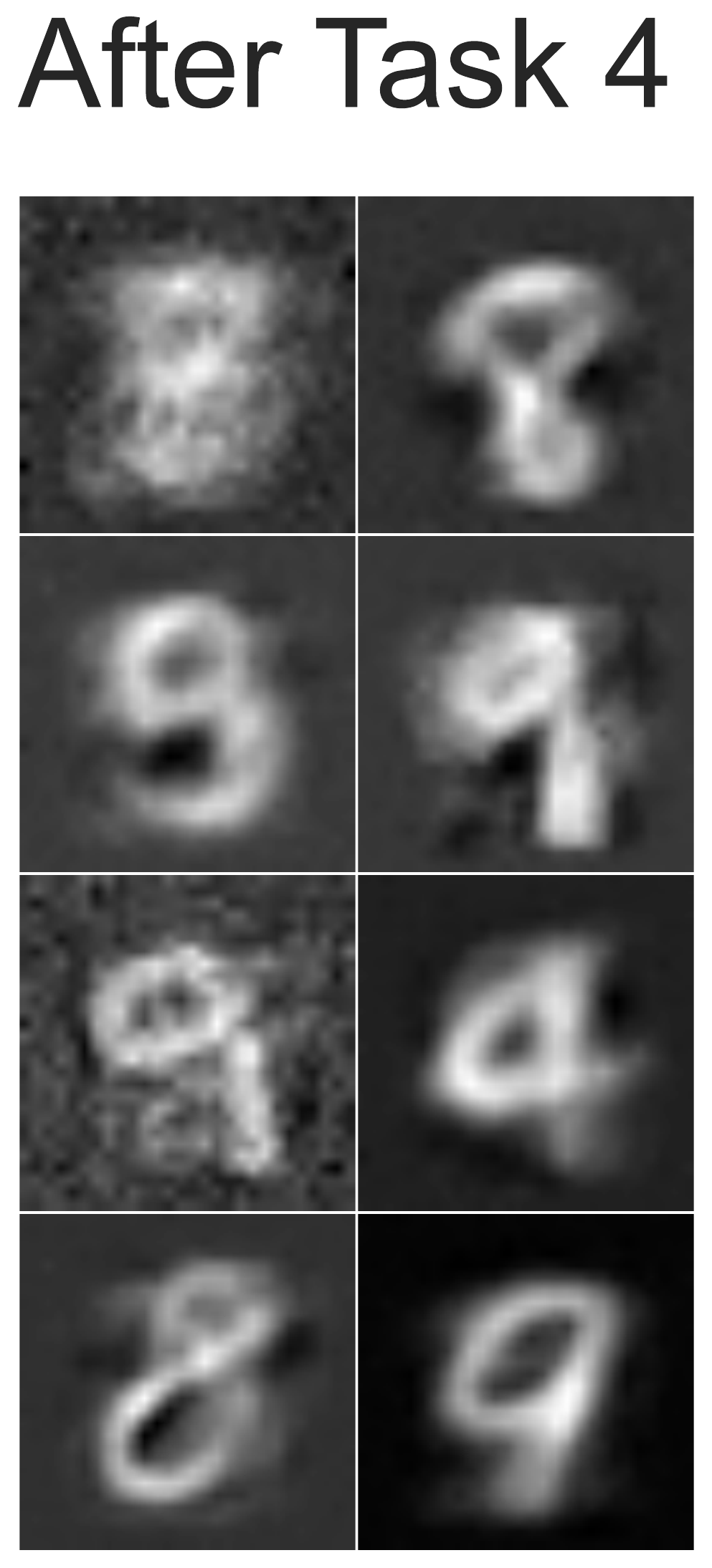}
\end{tabular}
\caption{
We visualize the inducing points learned after each task in Split MNIST,
progressing from left to right. This figure reveals that VAR-GPs spread inducing
points to cover the input space for each task.
}
\label{fig:viz_smnist_ind_pts}
\end{figure}
\egroup

\subsection{Benchmark Results}

With a qualitative picture of how VAR-GPs learn favorably for sequential tasks 
in continual learning, we validate its performance on the benchmark datasets of 
Split MNIST and Permuted MNIST.

\begin{figure}[!ht]
\centering
\begin{tabular}{cc}
\includegraphics[width=.45\linewidth]{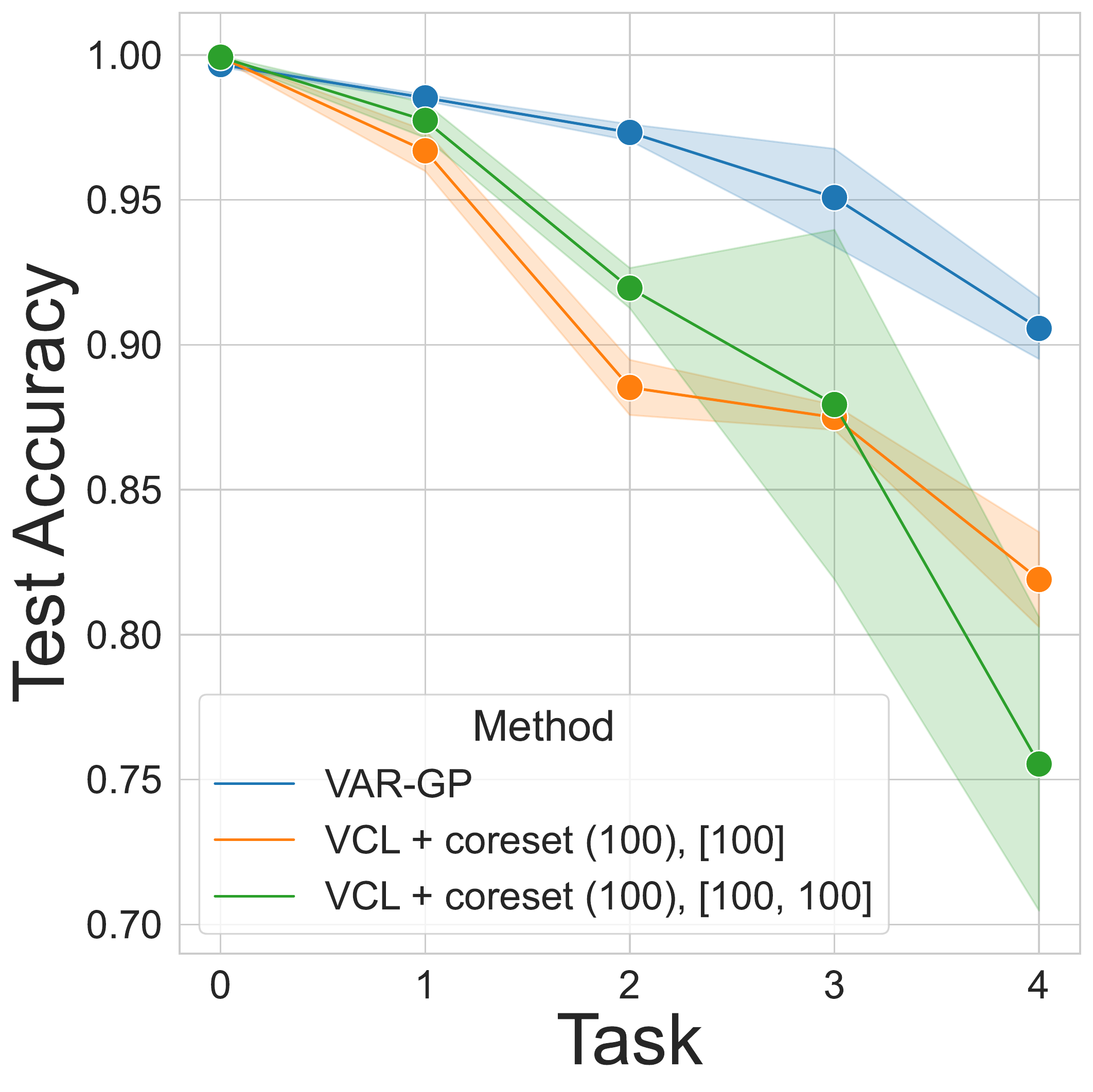} &
\includegraphics[width=0.45\linewidth]{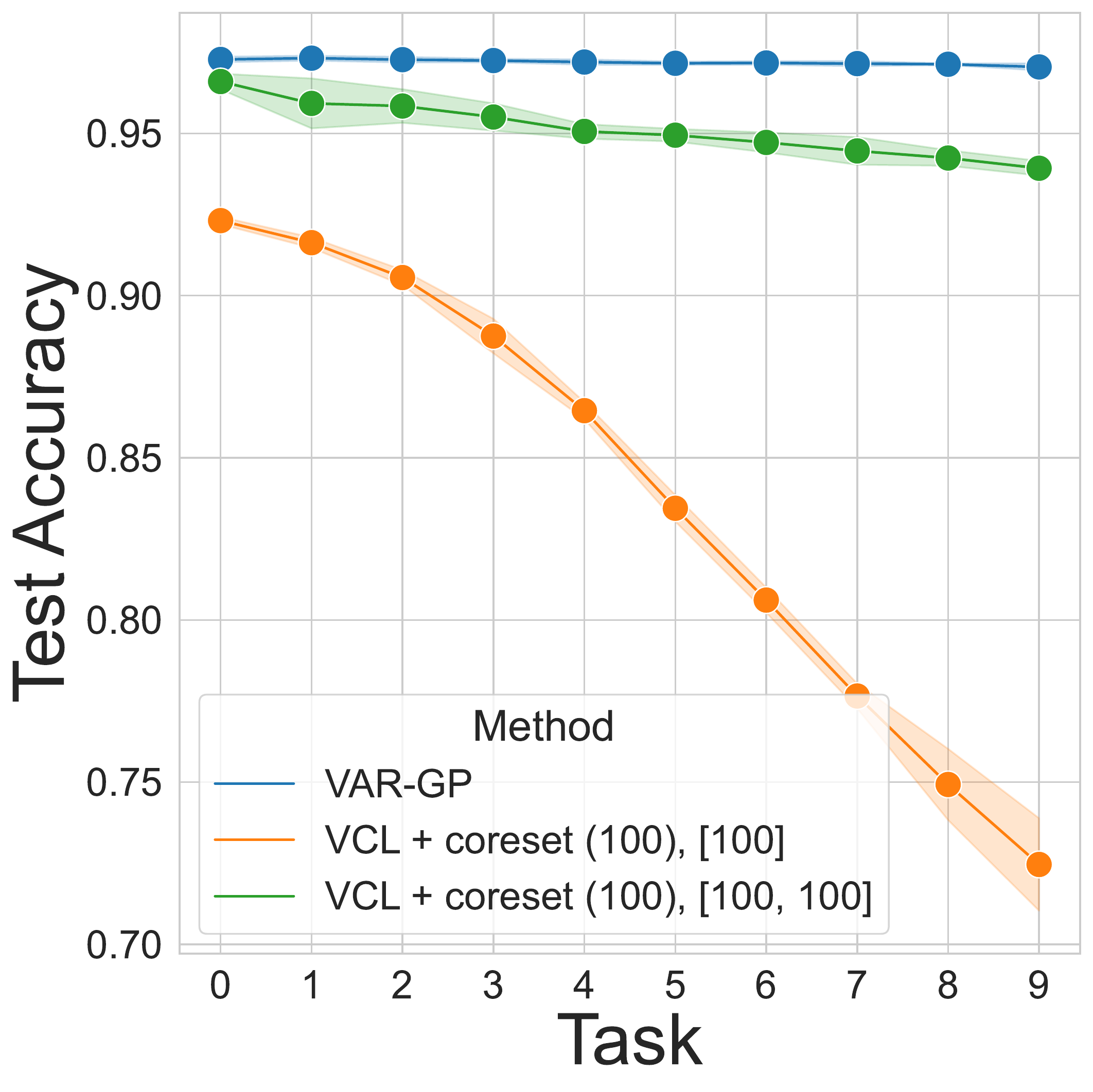} \\
(a) Split MNIST & (b) Permuted MNIST
\end{tabular}
\caption{
Comparing VAR-GPs to variants of VCL \citep{nguyen2017variational, 
swaroop2019improving} as we see more tasks (shown along the X-axis). We compute 
the average test accuracy (show along Y-axis) across all tasks seen so far, 
including the current one. The plots show mean and one standard deviation across 
five independent trials. VAR-GPs preserve information from old tasks, avoid 
catastrophic forgetting and consequently perform much better on newer tasks 
without compromising on old ones. See \cref{tab:results}, for all numerical 
results.
}
\label{fig:split_perm_mnist}
\end{figure}

\cref{fig:split_perm_mnist} shows how the cumulative performance across all
tasks seen so far evolves as the number of tasks increase. VAR-GPs perform 
favorably by performing consistently better throughout. One of the most 
competitive methods, VCL and its variants involving episodic memory through 
coresets \citep{swaroop2019improving} tend to deteriorate much faster, hinting 
at catastrophic forgetting. Quantitatively, we find the final mean test accuracy 
achieved by VAR-GPs after training on five Split MNIST tasks is $90.57\%$, which 
is significantly better than the best performing VCL variant at $81.90\%$. 
Similarly, VAR-GPs achieve a mean test accuracy of $97.2\%$ after training on 
ten tasks of Permuted MNIST, as compared to $95.06\%$ by the best variant of 
VCL. Comparisons against other continual learning algorithms are summarized in 
\cref{tab:results}.

\begin{table}[!ht]
\centering
\caption{
The final average test accuracy (in $\%$) after sequential training on benchmark 
dataset tasks \textemdash five for Split MNIST and ten for 
Permuted MNIST are noted below. We provide the mean and one 
standard deviation (where available) over five independent trials. $^\star$Split 
MNIST results for SI \citep{zenke2017continual} and EWC 
\citep{kirkpatrick2017overcoming} are taken from VCL 
\citep{swaroop2019improving}, however, are not directly comparable as they use a 
multi-head setup making evaluation easier; CS = coreset.; $[100]$ and 
$[100,100]$ represent hidden layer sizes of the neural networks.}
\vspace{1em}
\label{tab:results}
\begin{adjustbox}{width=\linewidth}
\begin{tabular}{l|c|c}
\toprule
\textbf{Method} & \textbf{Split MNIST} & \textbf{Permuted MNIST} \\\midrule
SI  & 98.9$^\star$ & 86.02 \\
EWC & 63.1$^\star$ & 84.11 \\\midrule
VCL, [100] & 19.90 {\footnotesize $\pm$ 0.14} & 82.94 {\footnotesize $\pm$ 0.85} \\
\-\hspace{3mm} + CS(50) & 76.21 {\footnotesize $\pm$ 2.02} & 85.80 {\footnotesize $\pm$ 0.41} \\
\-\hspace{3mm} + CS(100) & 81.90 {\footnotesize $\pm$ 1.64} & 86.45 {\footnotesize $\pm$ 0.26} \\
VCL, [100, 100] & 19.91 {\footnotesize $\pm$ 7.79} & 94.31 {\footnotesize $\pm$ 1.05} \\
\-\hspace{3mm} + CS(50) & 71.89 {\footnotesize $\pm$ 5.06} & {\footnotesize 95.43 $\pm$ 0.39} \\
\-\hspace{3mm} + CS(100) & 75.54 {\footnotesize $\pm$ 1.06} & 95.06 {\footnotesize $\pm$ 0.22} \\\midrule
VAR-GP (ours) & \textbf{90.57} {\footnotesize $\pm$ 1.06} & \textbf{97.20} {\footnotesize $\pm$ 0.08} \\
\-\hspace{3mm} + Block Diag. & 78.64 {\footnotesize $\pm$ 1.41} & 96.31 {\footnotesize $\pm$ 0.42} \\
\-\hspace{3mm} + MLE Hypers & 10.09 {\footnotesize $\pm$ 0.40} & 10.07 \footnotesize{$\pm$ 0.15} \\
\-\hspace{3mm} + Global & 39.31 {\footnotesize$\pm$ 0.28} & 46.02 {\footnotesize $\pm$ 1.09} \\\bottomrule
\end{tabular}
\end{adjustbox}
\end{table}

\subsection{Ablations} \label{sec:ablations}

Finally, we validate the specific modeling choices in VAR-GPs via a thorough 
ablation study.

\begin{figure}[!ht]
\centering
\begin{tabular}{cc}
\includegraphics[width=0.45\linewidth]{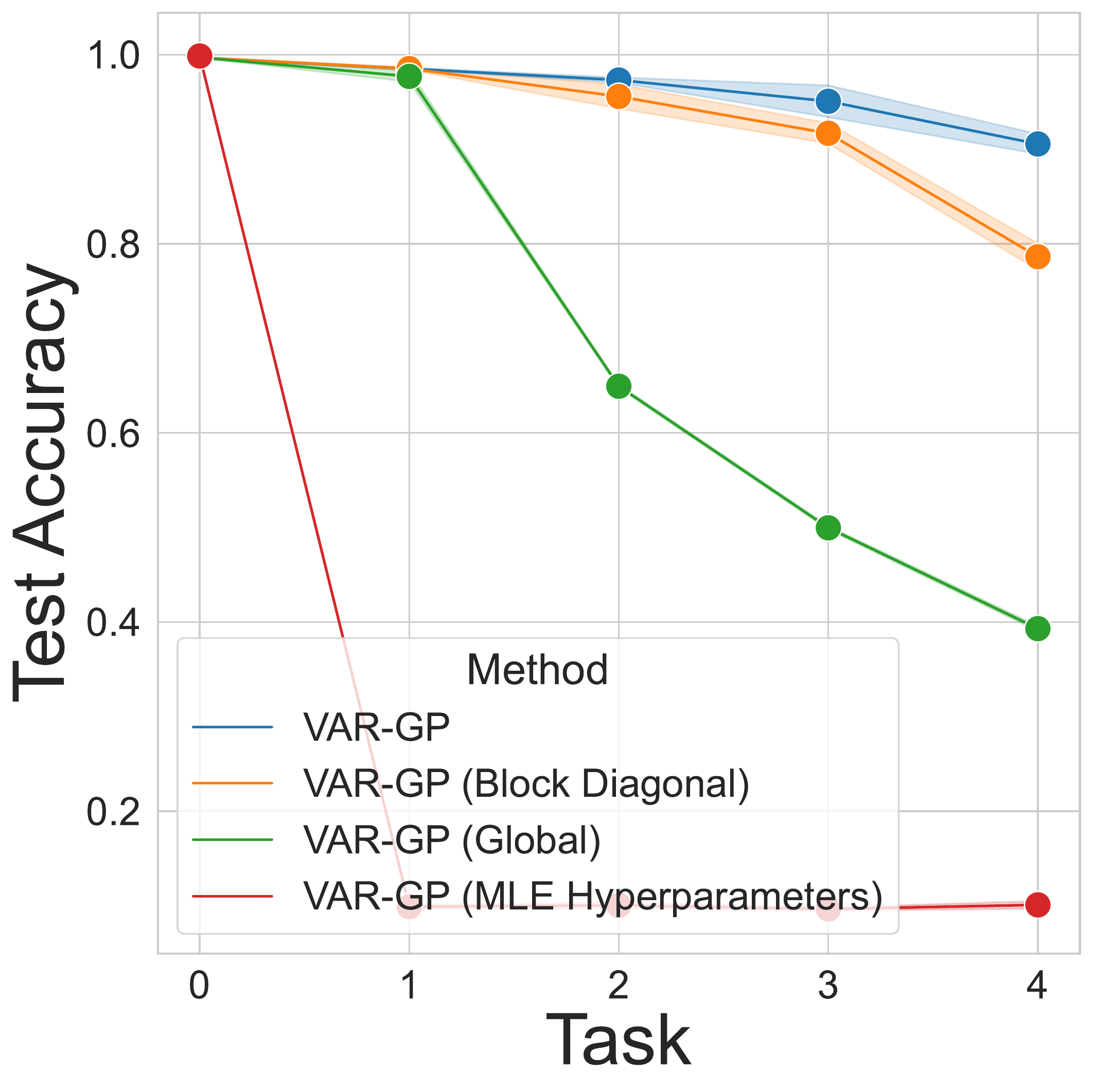} & 
\includegraphics[width=0.45\linewidth]{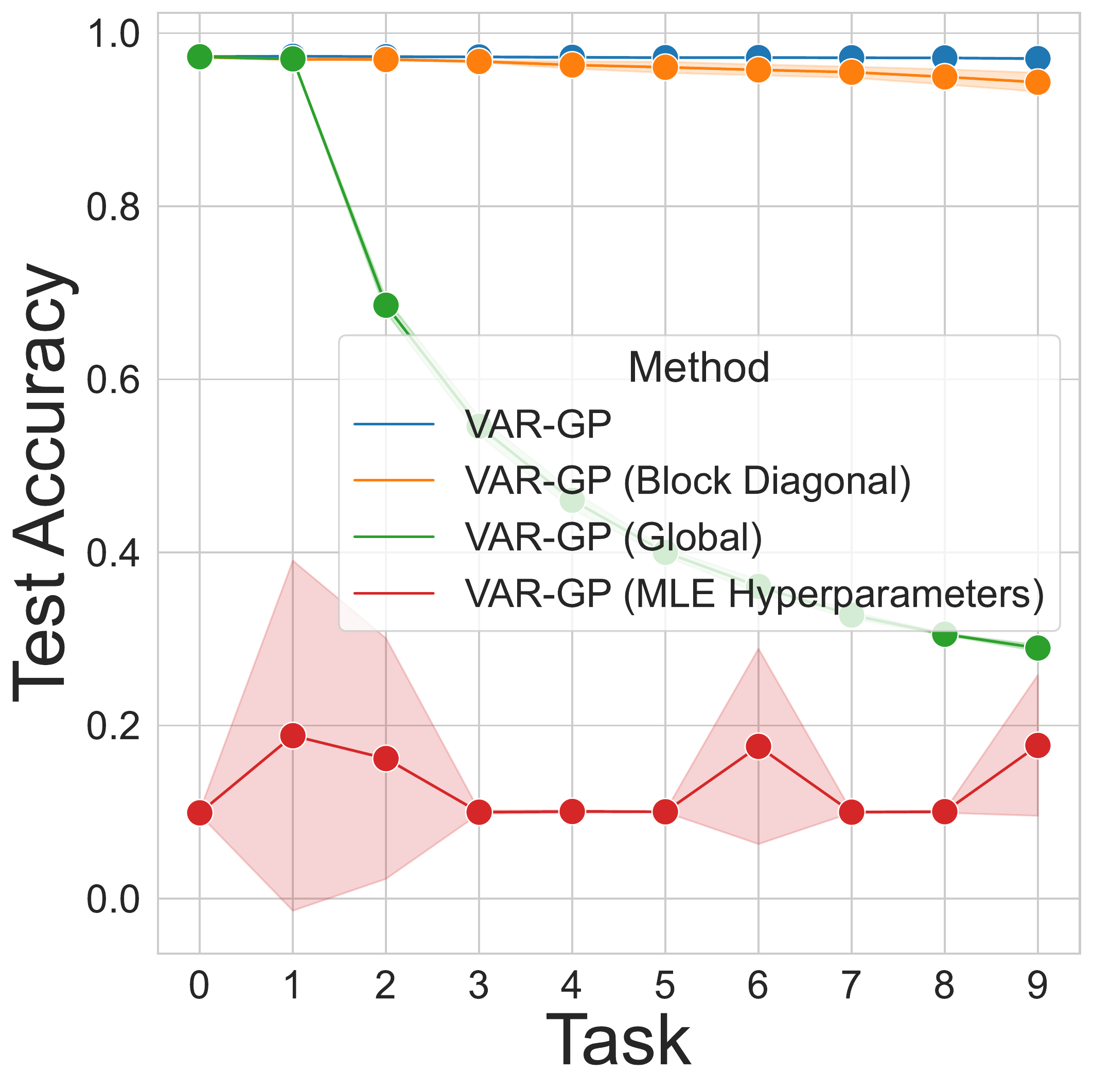} 
\\
(a) Split MNIST & (b) Permuted MNIST
\end{tabular}
\caption{
A thorough ablation study reveals the significance of our modeling choices. 
Evidently, VAR-GPs are able to sustain long-term continual learning performance 
than any other variant. See \cref{sec:ablations} for a detailed description of 
the ablations conducted.
}
\label{fig:split_perm_mnist_ablation}
\end{figure}

\paragraph{Block-Diagonal Variational Distribution} Instead of the 
auto-regressive posterior ${q(\uvec_t \mid \uvec_{< t}, \hpvec) = 
\mathcal{N}(\uvec_t; \Kvec_{\Zvec_t, \Zvec_{< t}} \Kvec_{\Zvec_{< t}, 
\Zvec_{< t}}^{-1} \uvec_{< t} + \mvec_t, \Svec_t)}$, we choose a simpler 
variational distribution given by ${q^{\prime}(\uvec_t \mid \uvec_{< t}, \hpvec) 
= \mathcal{N}(\uvec_t; \mvec_t, \Svec_t)}$, keeping everything else the same. 
Effectively, removing the conditioning induces a block diagonal structure in the 
full covariance matrix across the inducing variables where each block represents 
the covariance among the inducing points for a given task. This also decouples 
the inducing points and the hyperparameters in the approximate posterior. We 
hypothesize that this is detrimental to performance. 
\cref{fig:split_perm_mnist_ablation} shows that the long-term continual learning 
performance tends to deteriorate faster, hinting at \emph{greater} catastrophic 
forgetting.

\paragraph{Global Inducing Points} In an alternative model, we completely do
away with the auto-regressive nature of the variational distribution and just 
rely on a single set of inducing points at each time step 
${q_t(f, \hpvec) =
q_t(\hpvec) p(f_{\neq\uvec_{t}} \mid \uvec_{t}, \hpvec) q(\uvec_{t})}$.
\citet{bui2017streaming} rely on this variational distribution for the inducing 
points and an MLE estimate of the hyperparameters for streaming regression 
tasks. See \cref{sec:app_global_ind_pts} for precise modeling details and 
derivations. The experimental evidence in \cref{fig:split_perm_mnist_ablation}, 
however, shows that such variational approximation is poor for large scale 
continual classification tasks.

\paragraph{MLE Hyperparameters} Quantifying uncertainty about the 
hyperparameters considerably helps the model to perform well across tasks 
without a detrimental effect on the old ones. In this section, we validate this 
hypothesis by simply switching off the $\kl$-divergence term for the 
hyperparameters in both \cref{eq:elbo_1,eq:elbo_general}. Instead, we rely on a 
point estimate of the hyperparameters and use the maximum likelihood estimate at 
each step. The stark performance comparison is shown in 
\cref{fig:split_perm_mnist_ablation}. The hyper-parameters are stuck in a local 
minimum and virtually never recover for subsequent tasks.

\paragraph{Retraining Old Inducing Points} We also investigate a variant of 
VAR-GP named Re-VAR-GP where unlike earlier, we retrain the old inducing points 
$\mathbf{Z}_{<t}$. This changes the form of the variational lower bound. The 
precise details and implications are discussed in \cref{sec:app_revargp}, owing 
to which we do not pursue this approach further.

\section{Discussion} \label{sec:summary}

In this work, we develop VAR-GPs, a principled Bayesian inference scheme for 
continual learning. Using a sparse inducing point approximation, we propose a 
structured variational approximation to the true posterior for Gaussian 
processes, which reveals two fruitful connections to Expectation Propagation and 
Orthogonal Inducing Points. The resulting lower bound provides a natural 
learning objective to update the belief over underlying functions as new data 
arrives. A qualitative characterization of how VAR-GPs learn is presented, 
backed by strong empirical results on modern continual learning benchmarks. 
Further, a thorough ablation study establishes the efficacy of modeling choices.

In its current form, the number of inducing points grows linearly in the number 
of tasks. Consequently, the inference grows cubic in the number of tasks. This 
computational complexity limits the scaling of VAR-GPs to only a moderate number 
of continual learning tasks. Nevertheless, accommodating a post-hoc information 
distillation procedure in the learning process can prove helpful for 
scalability, and remains an open research problem. While Gaussian processes 
provide a flexible approach towards priors in the functional space, the 
representational power of VAR-GPs can be improved by the use of deep kernel 
learning. This, however, leads to new regularization challenges to be tackled 
(see \cref{sec:app_dkl}). Contemporary approaches relying on a memory buffer are 
also amenable to the proposed inference scheme, since the inducing points 
provide an automated data selection procedure. Finally, model-based 
reinforcement learning, where dataset shift is unavoidable, is a promising 
application space for VAR-GPs.

Through VAR-GPs, we remain optimistic that Gaussian processes provide an 
effective foundation for predictions under uncertainty in continual learning. 
More broadly, the characteristics highlighted by our learning objective 
\cref{eq:elbo_general} serve as a principled starting point to guide future 
research in continual learning objectives.

\section*{Acknowledgements}

We would like to thank Matthias Poloczek for insightful conversations along the 
way. SK was supported by the Uber AI Residency program.

\bibliography{references}
\bibliographystyle{icml2021}

\clearpage
\appendix

\twocolumn[
\icmltitle{Appendix for \\ Variational Auto-Regressive Gaussian Processes for Continual Learning}

\icmlsetsymbol{equal}{*}

\begin{icmlauthorlist}
\icmlauthor{Sanyam Kapoor}{nyu}
\icmlauthor{Theofanis Karaletsos}{fb}
\icmlauthor{Thang D. Bui}{usyd}
\end{icmlauthorlist}

\icmlaffiliation{nyu}{Center for Data Science, New York University, New York, NY, USA}
\icmlaffiliation{fb}{Facebook Inc., Menlo Park, CA, USA}
\icmlaffiliation{usyd}{University of Sydney, Sydney, NSW, Australia}

\icmlcorrespondingauthor{Sanyam Kapoor}{sanyam@nyu.edu}

\icmlkeywords{Gaussian processes, continual learning, variational inference}

\vskip 0.3in
]


\section{VAR-GPs}

\subsection{Posterior Predictive}

For a novel input $\xvec_{\star}$, the posterior predictive is computed via a 
Monte Carlo approximation of
\begin{align}
p(y_{\star} \mid \xvec_{\star} ) = \int p(y_{\star} \mid f) q_t(f, \hpvec \mid \xvec_{\star}) df d\hpvec
\end{align}

For a $K$-way classifier, we train $K$ independent GPs and use the Bayes optimal 
prediction ${\argmax_{i} p(y_{\star}^{(i)} \mid \xvec_{\star})}$, for all 
${i \in \{ 1, \dots, K \}}$, to compute accuracies.

\section{Ablations}

\subsection{Global Inducing Points} \label{sec:app_global_ind_pts}

This section outlines the assumptions made for the ablation titled as ``Global". 
The characterization for the first task remains the same as in VAR-GPs. For 
subsequent tasks, the general model and the variational assumption is written as 
(with implicit dependence on $\Zvec$),
\begin{align}
\begin{split}
p(\yvec^{(t)}, f, \hpvec \mid &~ \Xvec^{(t)}, \mathcal{D}^{(< t)}) \approx \\
&~\prod_{i=1}^{N_t} p(y_i^{(t)} \mid f, \xvec_i^{(t)}) \\
&~ p(f_{\neq \uvec_{t - 1}} \mid \Xvec^{(t)}, \uvec_{t-1}, \hpvec) \\
&~ q(\uvec_{t-1})q_{t-1}(\hpvec) ~.
\end{split}
\end{align}

Note that we don't have the auto-regressive characterization of 
VAR-GPs in the model anymore and instead have an approximate dependence through 
the variational posterior for the previous task. We further note that,
\begin{align}
p(f_{\neq \uvec_{t-1}} \big| \Xvec^{(t)}, \uvec_{t-1}, \hpvec) =&{}~p(f_{\neq \uvec_{t-1}, \uvec_{t}} \big| \Xvec^{(t)}, \uvec_{t-1}, \uvec_{t}, \hpvec) \nonumber \\
&{} \frac{p(\uvec_{t-1},\uvec_t \big | \hpvec)}{p(\uvec_{t-1} \big| \hpvec)}, \\
p(f_{\neq \uvec_{t}} \big| \Xvec^{(t)}, \uvec_{t}, \hpvec) =&{} ~p(f_{\neq \uvec_{t-1}, \uvec_{t}} \big| \Xvec^{(t)}, \uvec_{t-1}, \uvec_{t}, \hpvec) \nonumber \\
&{} \frac{p(\uvec_{t-1},\uvec_t | \hpvec)}{p(\uvec_t \big| \hpvec)}.
\end{align}

Owing to key cancellations, the variational lower bound now is given by,

\begin{align}
\begin{split}
\mathcal{F}(&q_t) = \sum_{i=1}^{N_t} \mathbb{E}_{q_t(f,\hpvec)}\left[ \log{p\left(y_i^{(t)} \mid f, \xvec_i^{(t)} \right)} \right] \\
&- \kl\left[ q_t(\hpvec) \mid\mid q_{t-1}(\hpvec) \right] \\
&- \mathbb{E}_{q_t(\hpvec)}\left[ \kl\left[ q(\uvec_{t}) \mid\mid p(\uvec_{t} \mid \hpvec) \right] \right] \\
&+ \mathbb{E}_{q_t(\hpvec)q(\uvec_{t} \mid \hpvec)p(\uvec_{t-1} \mid \uvec_t, \hpvec)} \left[ \log{ \frac{q(\uvec_{t-1})}{p(\uvec_{t-1} \mid \hpvec)} } \right]~.
\end{split}
\end{align}

A key difference to note here is that the $\kl$ regularizer containing inducing 
points $\uvec_t$ is not conditional on the previous ones anymore.

\subsection{Re-VAR-GP: Retraining old inducing points}\label{sec:app_revargp}

In this version of VAR-GP, we allow retraining of old inducing points and call 
it Retrainable VAR-GP (abbreviated as Re-VAR-GP). We clarify the precise nature 
of terms. Leading from \cref{eq:general_cont_model,eq:elbo_general}, we 
highlight frozen $\widetilde{\uvec}_{<t}$ and $\widetilde{\Zvec}_{<t}$ (with a 
tilde) in the prior model to differentiate against learnable parameters,
\begin{align}
\begin{split}
p(\yvec^{(t)}, f, \hpvec &~\mid  \Xvec^{(t)}, \mathcal{D}^{(< t)}) = \prod_{i=1}^{N_t} p(y_i^{(t)} \mid f, \xvec_i^{(t)}) \\
&~ p(f_{\neq \uvec_t, \widetilde{\uvec}_{<t}} \mid \Xvec^{(t)}, \uvec_t, \widetilde{\uvec}_{<t}, \Zvec_{t}, \widetilde{\Zvec}_{< t}, \hpvec) \\
&~ p(\uvec_t \mid \Zvec_t, \widetilde{\uvec}_{< t}, \widetilde{\Zvec}_{< t}, \hpvec) \\
&~ q(\widetilde{\uvec}_{< t} \mid \widetilde{\Zvec}_{< t}, \hpvec) q_{t-1}(\hpvec)~.
\end{split}	
\end{align}

We posit the variational posterior as,
\begin{align}
\begin{split}
q_t(f, \hpvec) =&~ p(f_{\neq \uvec_t, \uvec_{<t}} \mid \Xvec^{(t)}, \uvec_t, \uvec_{<t}, \Zvec_{t}, \Zvec_{< t}, \hpvec) \\
&~ q(\uvec_t \mid \Zvec_t, \uvec_{< t}, \Zvec_{< t}, \hpvec) \\
&~ q(\uvec_{< t} \mid \Zvec_{< t}, \hpvec)q_{t}(\hpvec)
\end{split}	
\end{align}

To simplify these equations, we note the following identities.
\begin{align*}
\begin{split}
&p(f_{\neq \uvec_t, \uvec_{<t}} \mid \Xvec^{(t)}, \uvec_t, \uvec_{<t}, \Zvec_{t}, \Zvec_{< t}, \hpvec)	 = \\
&p(f_{\neq \uvec_t, \uvec_{<t}, \tilde{\uvec}_{<t}} \mid \Xvec^{(t)}, \uvec_t, \uvec_{<t}, \tilde{\uvec}_{<t}, \Zvec_{t}, \Zvec_{< t}, \tilde{\Zvec}_{< t}, \hpvec) \\
&p(\tilde{\uvec}_{<t} \mid \uvec_t, \uvec_{<t}, \Zvec_{t}, \Zvec_{<t}, \tilde{\Zvec}_{<t}, \hpvec)~,
\end{split}	
\end{align*}

\begin{align*}
\begin{split}
&p(f_{\neq \uvec_t, \tilde{\uvec}_{<t}} \mid \Xvec^{(t)}, \uvec_t, \tilde{\uvec}_{<t}, \Zvec_{t}, \tilde{\Zvec}_{< t}, \hpvec) = \\
&p(f_{\neq \uvec_t, \uvec_{<t}, \tilde{\uvec}_{<t}} \mid \Xvec^{(t)}, \uvec_t, \uvec_{<t}, \tilde{\uvec}_{<t}, \Zvec_{t}, \Zvec_{< t}, \tilde{\Zvec}_{< t}, \hpvec) \\
&p(\uvec_{<t} \mid \uvec_t, \tilde{\uvec}_{<t}, \Zvec_{t}, \Zvec_{<t}, \tilde{\Zvec}_{<t}, \hpvec)~,	
\end{split}	
\end{align*}

\begin{align*}
\begin{split}
&\frac{p(\uvec_t \mid \Zvec_t, \tilde{\uvec}_{< t}, \tilde{\Zvec}_{< t}, \hpvec)p(\uvec_{<t} \mid \uvec_t, \tilde{\uvec}_{<t}, \Zvec_{t}, \Zvec_{<t}, \tilde{\Zvec}_{<t}, \hpvec)}{p(\tilde{\uvec}_{<t} \mid \uvec_t, \uvec_{<t}, \Zvec_{t}, \Zvec_{<t}, \tilde{\Zvec}_{<t}, \hpvec)} \\
&=\frac{p(\uvec_{<t}, \uvec_t \mid \Zvec_t, \Zvec_{<t}, \hpvec)}{p(\tilde{\uvec}_{<t} \mid \tilde{\Zvec}_{<t},\hpvec)}.	
\end{split}	
\end{align*}

Using these identities, the lower bound now simplifies as,
\begin{footnotesize}
\begin{align}
\begin{split}
\mathcal{F}(q_t) =&~ \sum_{i=1}^{N_t} \mathbb{E}_{q_t(f,\hpvec)}\left[ \log{p(y_i^{(t)} \mid f, \xvec_i^{(t)} )} \right]	 \\
& - \kl\left[ q_t(\hpvec) \mid\mid q_{t-1}(\hpvec) \right] \\
& - \mathbb{E}_{q_t(\hpvec)} \left[ \kl\left[ q(\uvec_{\leq t} \mid \Zvec_{\leq t}, \hpvec) \mid\mid p(\uvec_{\leq t} \mid \Zvec_{\leq t}, \hpvec) \right] \right] \\
& - \mathbb{E}_{q_t(\hpvec)q(\uvec_{\leq t} | \Zvec_{\leq t}, \theta)p(\tilde{\uvec}_{<t} | \tilde{\Zvec}_{< t}, \uvec_{\leq t}, \Zvec_{\leq t}, \hpvec)}\left[\mathfrak{R}_t \right]~,
\end{split} \label{eq:elbo_vargp_retrain}
\end{align}	
\end{footnotesize}

where ${\mathfrak{R}_t = \log{
\frac{p(\tilde{\uvec}_{<t}| \tilde{\Zvec}_{<t}, \hpvec)}
{q( \tilde{\uvec}_{<t} | \tilde{\Zvec}_{<t}, \hpvec)}}}$.
The key differences to note here are the fact that prior model now conditions on 
the frozen inducing points while the new variational distributions introduced 
are still free to optimize those points further. This leads to additional terms 
in the variational lower bound. We briefly discuss the performance next.

\subsubsection{Performance on Toy Dataset}

\begin{figure}[!ht]
\centering
\includegraphics[width=.95\linewidth]{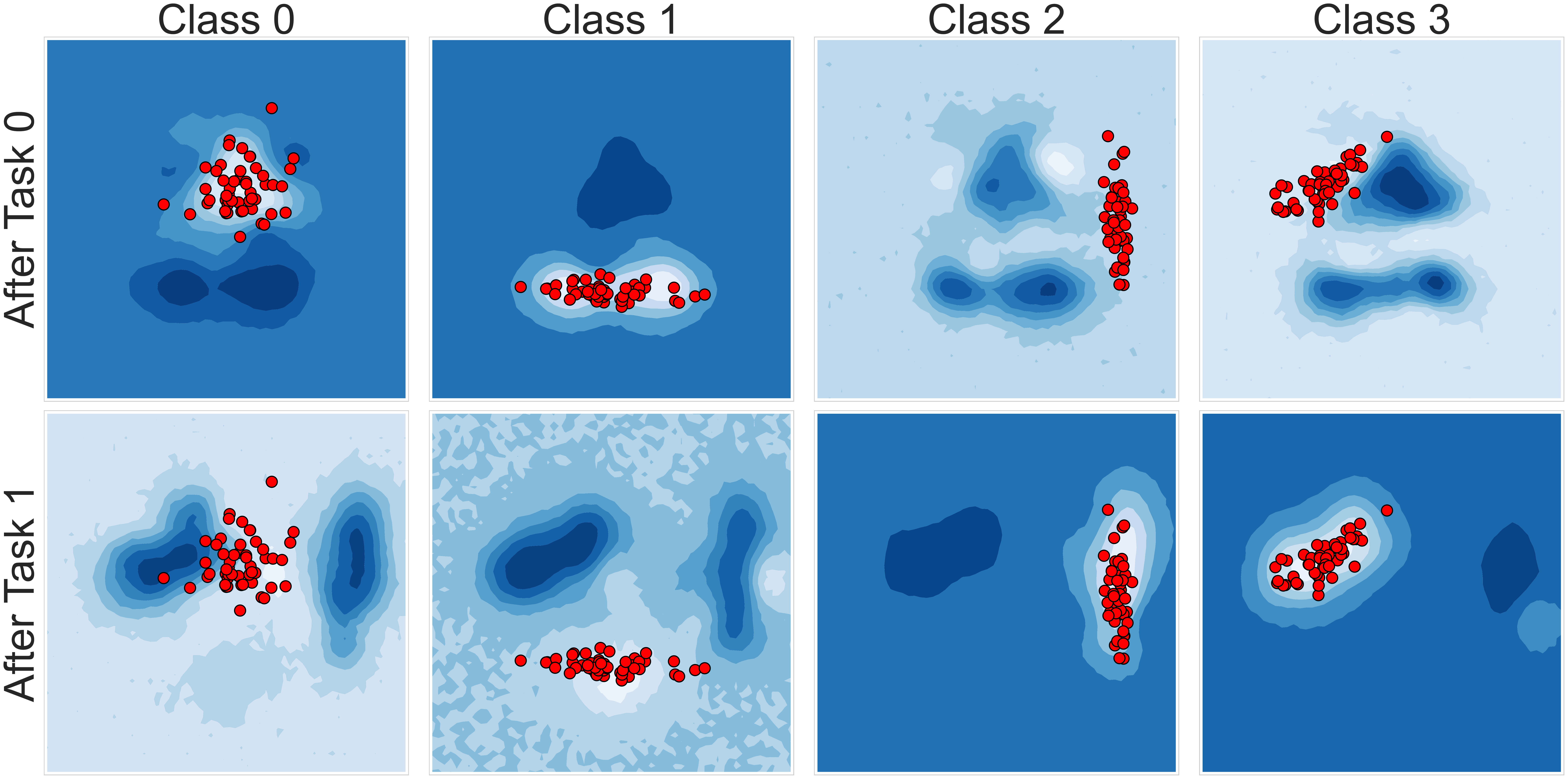}
\caption{
This figure shows class-wise output probabilities in each column for classifiers 
trained using a synthetic dataset (\cref{fig:toy_data}) on the 2-D plane 
$x,y \in [-3., 3.]$. The first row represents the density surface after training 
for \textbf{Task 0} (observing classes $0/1$) and the second after training for 
\textbf{Task 1} (observing classes $2/3$). Brighter regions represent higher 
probabilities. Training points for each class are marked 
\tikzcircle[fill=red]{2.5pt}. Re-VAR-GP tends to suffer from catastrophic 
forgetting. Notice the approximately uniform uncertainty in regions for Class $0$
and Class $1$ after training on the second task.
}
\label{fig:toy_2d_retrain}
\end{figure}

Similar in spirit to \cref{fig:toy_2d}, we train Re-VAR-GP on the toy dataset in 
\cref{fig:toy_data}. The density plots for training after both first task 
(training on classes $0/1$) and the second (training on classes $2/3$) are 
presented in \cref{fig:toy_2d_retrain}.

As shown in \cref{fig:toy_2d_retrain}, Re-VAR-GP is not able to retain the 
information gained from previous task, a sign of catastrophic forgetting. This 
can be understood from the nature of the lower bound in 
\cref{eq:elbo_vargp_retrain}. The only term that can potentially contribute to 
preservation of old information is the expected ratio 
${\log{ \frac{p(\tilde{\uvec}_{<t} \mid \tilde{\Zvec}_{<t}, \hpvec)}
{q( \tilde{\uvec}_{<t} \mid \widetilde{\Zvec}_{<t}, \hpvec)}}}$. However, this 
term avoids any interaction between $\widetilde{\uvec}_{\leq t}$ and 
$\uvec_{\leq t}$. As a result, the retrainable parameters $\uvec_{\leq t}$ and 
$\Zvec_{\leq t}$ have no information-preserving regularization unlike VAR-GPs as 
seen in \cref{eq:elbo_general}. Owing to this observation, we do not pursue this 
model further.

\subsection{Deep Kernel Learning} \label{sec:app_dkl}

For increased representational power in the kernel, we also provide preliminary 
experiments with Deep Kernel Learning \citep{wilson2016deep}. Effectively, we 
augment the Exponentiated Quadratic kernel with a feature extractor 
$g_{\phi}(x)$ in the form of a neural network and allow them to be trained with 
additional kernel hyperparameters $\phi$. This amounts to replacing $\xvec$ and 
$\xvec^\prime$ in \cref{eq:sqexpkern} with $g_\phi(\xvec)$ and 
$g_{\phi}(\xvec^\prime)$ respectively. We only use point estimates for $\phi$,
initialized at the previous task for all $t>1$.

\subsubsection{Experiments with Split MNIST}

\begin{figure}[!ht]
\centering
\includegraphics[width=.7\linewidth]{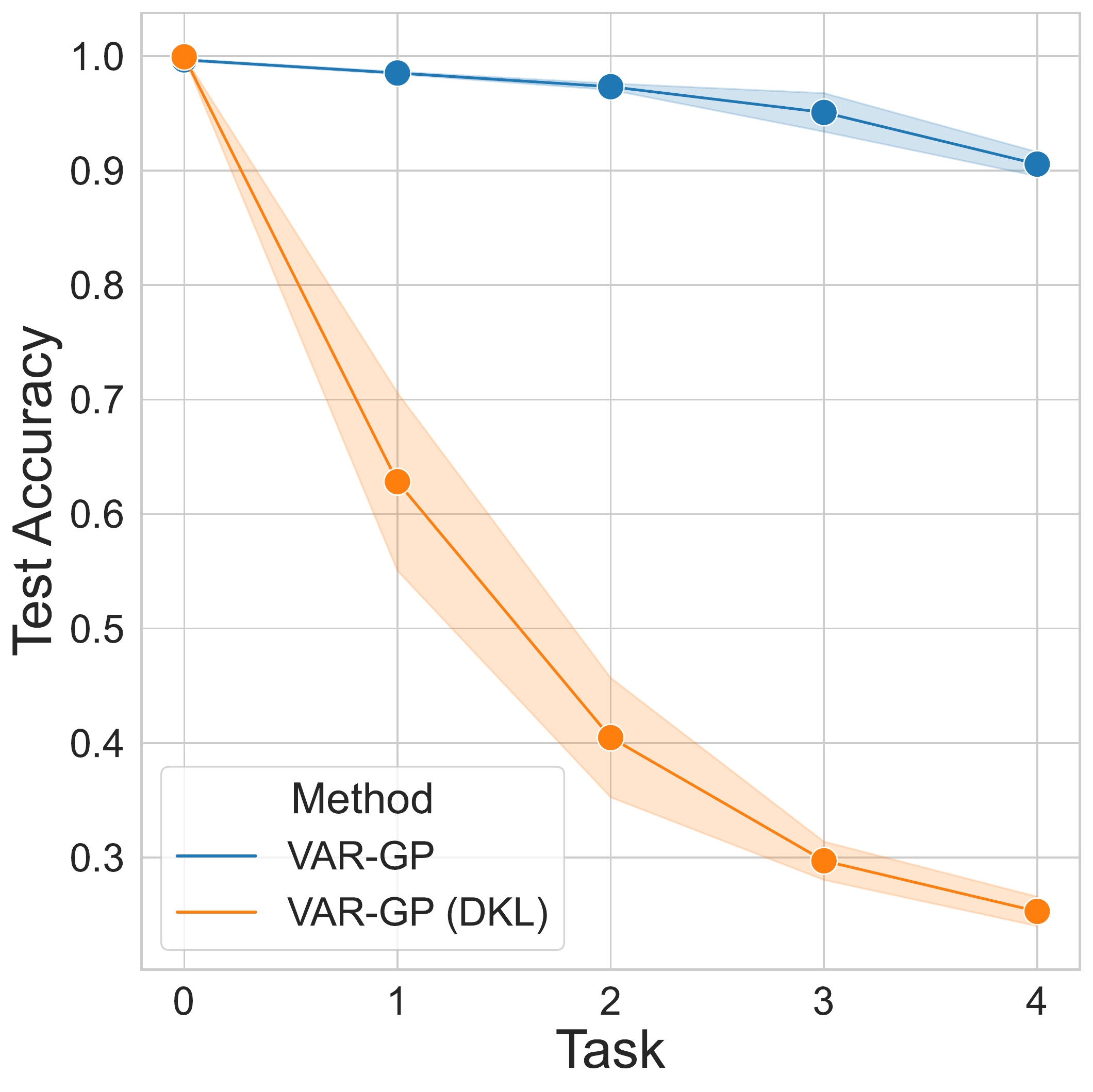}
\caption{
This figure shows the task-wise test accuracy on Split MNIST over five 
independent runs. We train a neural network as a feature extractor 
before applying the kernel (as described in \cref{sec:app_dkl}). It is 
clear that the neural networks require stronger regularization as we 
incorporate more tasks.
}
\label{fig:smnist_dkl}
\end{figure}

We use a neural network with two hidden layers of size $256$ each and the final 
output feature size of $64$ to parameterize $f_\phi$ and train the system 
end-to-end. As we see in \cref{fig:smnist_dkl}, the performance declines much 
faster than in VAR-GPs.

This result hints towards weak regularization of the feature extractor as we 
encounter subsequent tasks. The introduction of a neural network makes the 
inference problem much harder and any potential remedies are beyond the scope of 
current work. We, therefore, do not discuss this further but makes for an 
exciting direction to pursue in the future.

\section{Implementation}

\subsection{Exponentiated Quadratic kernel} \label{sec:sq_exp_kernel}

The precise parametrization of the kernel used is given below. 
$\norm{\cdot}_2$ is the $\ell^2$-norm. We parameterize $\log{\gamma}$ and 
$\log{\sigma}$.
\begin{align} \label{eq:sqexpkern}
k(\xvec, \xvec^\prime) = \gamma \exp{\left\{ \frac{|| \xvec - \xvec^\prime ||^2_2}{2 \sigma^2} \right\}}
\end{align}

\subsection{Parameterizing covariance matrices} \label{sec:parametrizing_cov}

In all experiments, we parameterize a covariance matrix 
${\Svec \in \mathbb{R}^{M \times M}}$ using its Cholesky decomposition 
${\Svec = \mathbf{L}\mathbf{L}^\intercal}$, where 
${\mathbf{L} \in \mathbb{R}^{M \times M}}$ is a lower triangular matrix with 
positive diagonals. The positivity of the diagonals is maintained via a 
\texttt{softplus} transform. As a result, we can apply unconstrained 
optimization on ${\tfrac{1}{2}(M \times (M+1))}$ free parameters corresponding
to the lower trianglular matrix $\mathbf{L}$.

\subsection{Computing the auto-regressive distributions in VAR-GPs}\label{sec:auto_regressive_trick}

When using the auto-regressive parametrization in VAR-GPs, the joint 
distribution over all inducing points up to and including the current time step 
can be decomposed as follows,
\begin{align}
\begin{split}
q(\uvec_{\leq t} | \hpvec) =	&~ q(\uvec_{< t} | \theta) q(\uvec_t | \uvec_{< t}, \hpvec) \\
=&~ \mathcal{N}(\uvec_{<t}; \mvec_{<t}, \Svec_{<t}) \\
&~\mathcal{N} (\uvec_{t};  \mathbf{A}_t \uvec_{< t} + \mvec_t, \Svec_t)~,
\end{split}	
\end{align}

such that ${\Avec_t = 
\Kvec_{\Zvec_t, \Zvec_{< t}}  \Kvec_{\Zvec_{< t}, \Zvec_{< t}}^{-1}}$.

While we cannot avoid sampling the hyperparameters $\hpvec$, we can avoid
variance introduced by the ancestral sampling of variational distribution for 
computation of \cref{eq:elbo_general}. We recognize that the full 
auto-regressive distribution can be computed in closed form as it is a product 
Gaussians with linear dependence in the mean, similar in spirit to linear 
Gaussian dynamical systems \citep{murphy2012machine}. Hence, for all $t>1$, we 
have
\begin{align}
q(\uvec_{\leq t}| \theta) = \mathcal{N}\left(\begin{bmatrix} \uvec_{< t} \\ \uvec_{t} \end{bmatrix}; \begin{bmatrix} \mvec_{< t} \\ \Avec_t \mvec_{< t} + \mvec_{t} \end{bmatrix}, \Svec \right)
\end{align}

where,
\begin{align*}
\Svec \defeq \begin{bmatrix} \Svec_{< t} & \Svec_{< t} \Avec_t^\intercal \\  \Avec_t \Svec_{< t}^\intercal & \Svec_t + \Avec_t \Svec_{< t} \Avec_t^\intercal \end{bmatrix}	
\end{align*}

\section{Hyperparameters}

\subsection{Search Space} \label{sec:hypers}

The search space for all hyperparameters used across experiments is described in 
\cref{tab:hp_searchspace}. Top hyperparameters were picked using a held-out 
validation set.

\begin{table}[!ht]
\caption{List of key hyperparameters with relevant search spaces.}
\label{tab:hp_searchspace}
\centering
\begin{tabular}{l | c}
\toprule
\textbf{Hyperparameter} & \textbf{Range / Value} \\\midrule
Learning Rate ($\eta$) & $[0.001, 0.01]$ \\
Inducing Points ($M$) & $[40, 200]$ \\
Hypers $\kl$ Tempering Factor ($\beta$) & $[1.0, 10.0]$ \\
Batch Size (B) & $512$ \\
Maximum Epochs (E) & $500$ \\
Early Stopping Patience Epochs (K) & $200$ \\
Early Stopping Tolerance ($\delta$) & $0.0001$ \\\bottomrule
\end{tabular}
\end{table}

\subsection{Varying number of inducing points M}

\begin{figure}[!ht]
\centering
\includegraphics[width=0.7\linewidth]{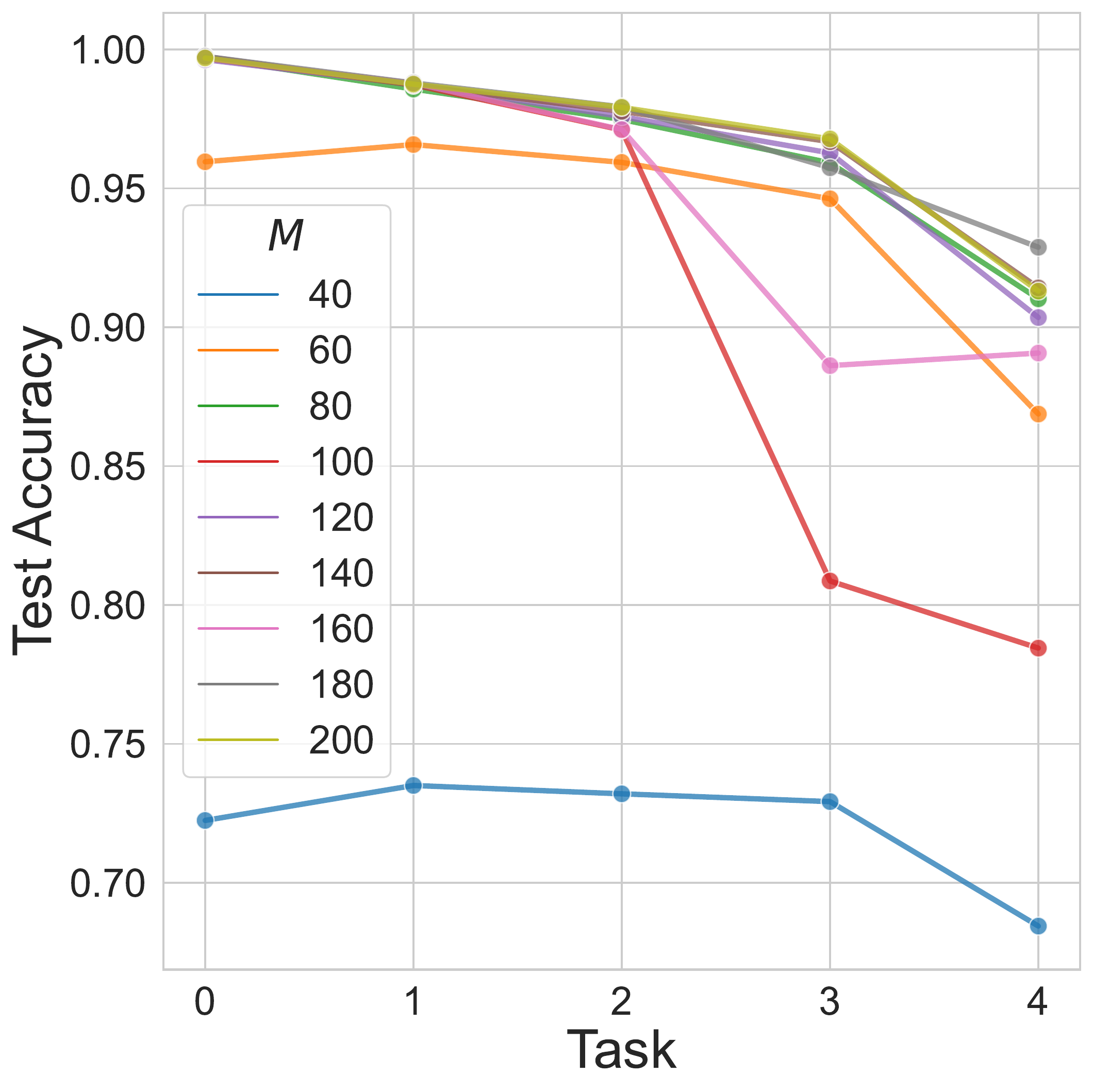}
\caption{
This figure shows the mean task-wise (on x-axis) test accuracy 
(y-axis) on Split MNIST over five independent runs, varying number of 
inducing points $M$ from $20$ to $200$ in steps of $20$. The key 
insight to draw here is about the general trend that increasing the 
number of inducing points $M$ improves the mean peformance, hinting 
towards more capacity being available to be exploited by the learning 
algorithm.
}
\label{fig:smnist_varying_M}
\end{figure}

In \cref{fig:smnist_varying_M}, we note the mean performance by varying the 
number of inducing points $M$ from $20$ to $200$ in steps of $20$, for Split 
MNIST. The key takeaway here is that increasing the number of inducing points 
generally does improve the performance. This indicates that there may be more 
capacity available to be exploited further.

\end{document}